%% file: root.tex
\documentclass[letterpaper, 10 pt, conference]{ieeeconf}

\IEEEoverridecommandlockouts                           

\usepackage{cite}
\usepackage{amsmath,amssymb,amsfonts}
\usepackage{wasysym}
\usepackage{graphicx}
\usepackage{textcomp}
\usepackage{xcolor}
\usepackage{multirow}
\usepackage{enumerate}
\usepackage{comment}
\usepackage[export]{adjustbox}
\usepackage[nolist,nohyperlinks]{acronym}
\usepackage{algorithm,algpseudocode}
\usepackage{url} 

\renewcommand{\baselinestretch}{0.848}
\DeclareUnicodeCharacter{2212}{\textendash}

\title{\LARGE \bf
Functional kinematic and kinetic requirements of the upper limb during activities of daily living: a recommendation on necessary joint capabilities for prosthetic arms}

\author{Christopher Herneth$^{1}$, Amartya Ganguly$^{1}$, and Sami Haddadin$^{1}$
\thanks{$^{1}$Christopher Herneth, Junnan Li, Muhammad Hilman Fatoni, Amartya Ganguly, and Sami Haddadin are with Chair of Robotics and Systems Intelligence, MIRMI - Munich Institute of Robotics and Machine Intelligence, Technical University of Munich (TUM), Munich, Germany {\tt\footnotesize \{christopher.herneth, junnan.li, mhilman.fatoni, amartya.ganguly, haddadin\}@tum.de}. © 2024 IEEE.  Personal use of this material is permitted.  Permission from IEEE must be obtained for all other uses, in any current or future media, including reprinting/republishing this material for advertising or promotional purposes, creating new collective works, for resale or redistribution to servers or lists, or reuse of any copyrighted component of this work in other works.} %
}

\begin{document}
\bibliographystyle{unsrt}

\maketitle
\thispagestyle{empty}
\pagestyle{empty}

\input{chapters/Abstract}

\setlength{\belowdisplayskip}{3pt} \setlength{\belowdisplayshortskip}{3pt}
\setlength{\abovedisplayskip}{3pt} \setlength{\abovedisplayshortskip}{3pt}

\input{chapters/Introduction}
\input{chapters/Methodology}

\input{chapters/Results}

\input{chapters/Discussion}
\input{chapters/Acronyms}

\section{Supplementary Material}
Parameters $K_{p,C}$ and coefficients of determination $R$ of individual \ac{LRM}s may be found here: \url{https://github.com/ChristopherHerneth/Supplementary-Material-Publications}. Additional video material may be accesses here: \url{https://youtu.be/WffbXRzqWqM}.

\section*{ACKNOWLEDGMENT}
This work was supported by the Federal Ministry of Education and Research of the Federal Republic of Germany (BMBF) by funding the project AI.D under the Project Number 16ME0539K.

\bibliography{Bibliography}

\end{document}

%% file: chapters/Abstract.tex
\begin{abstract}
Prosthetic limb abandonment remains an unsolved challenge as amputees consistently reject their devices. Current prosthetic designs often fail to balance human-like performance with acceptable device weight, highlighting the need for optimised designs tailored to modern tasks. This study aims to provide a comprehensive dataset of joint kinematics and kinetics essential for performing activities of daily living (ADL), thereby informing the design of more functional and user-friendly prosthetic devices. Functionally required Ranges of Motion (ROM), velocities, and torques for the Glenohumeral (rotation), elbow, Radioulnar, and wrist joints were computed using motion capture data from 12 subjects performing 24 ADLs. Our approach included the computation of joint torques for varying mass and inertia properties of the upper limb, while torques induced by the manipulation of experimental objects were considered by their interaction wrench with the subject's hand. 
Joint torques pertaining to individual ADL scaled linearly with limb and object mass and mass distribution, permitting their generalisation to not explicitly simulated limb and object dynamics with linear regressors (LRM), exhibiting coefficients of determination $R$ = 0.99 $\pm$ 0.01. 
Exemplifying an application of data-driven prosthesis design, we optimise wrist axes orientations for two serial and two differential joint configurations. Optimised axes reduced peak power requirements, between 22\% to 38\% compared to anatomical configurations, by exploiting high torque correlations ($r = -0.84, p < 0.05$) between Ulnar deviation and wrist flexion/extension joints. 
This study offers critical insights into the functional requirements of upper limb prostheses, providing a valuable foundation for data-driven prosthetic design that addresses key user concerns and enhances device adoption.
\end{abstract}

%% file: chapters/Introduction.tex
\section{Introduction}
Prosthetic limbs are crucial in improving lives and empowering individuals with limb amputations. However, despite extensive research, 23-50\% of prosthetic users do not use their devices consistently \cite{Espinosa2019_understandingAbbandonment, Biddiss2007_25Years_abbandonment_studyreview}. Prosthesis non-use and resulting unilateral performance may lead to cumulative trauma and severe injury of the remaining limb \cite{gambrellOveruseSyndromeUnilateral2008, ostlieMusculoskeletalPainOveruse2011, resnikAdvancedUpperLimb2012}, accentuating the need for user-friendly prosthetics. At the same time, 72\% of prosthesis non-users with limb differences would reconsider wearing a device if certain improvements were to be made \cite{Biddiss2007_25Years_abbandonment_studyreview}. 

Excessive device weight and poor weight distribution were identified as key concerns for users \cite{Davidson2002_Austr_Prost_Use, Biddiss2007_25Years_abbandonment_studyreview, Cordella2016_Need_ULPU}, ranking as the second most influential aspect for device abandonment after socket comfort \cite{Schultz2007_ExpertOpinnion_ULP_success}. Additional factors associated with high levels of dissatisfaction pertain to prosthesis agility, strength and control. Concurrently, upper limb prostheses were shown to be of limited effectiveness in basic tasks concerning independent living, work, household, and hobby \cite{Kerver2021_prefusagefeat, Biddiss2007_prosthconsumdesignfeat}, while grabbing, picking up, and holding were considered essential prosthesis capabilities by users \cite{Kerver2021_prefusagefeat}. 

\begin{figure}[t]
    \centering
    \fontsize{6.5pt}{6.5pt}\selectfont
    \def\svgwidth{0.5\textwidth}
    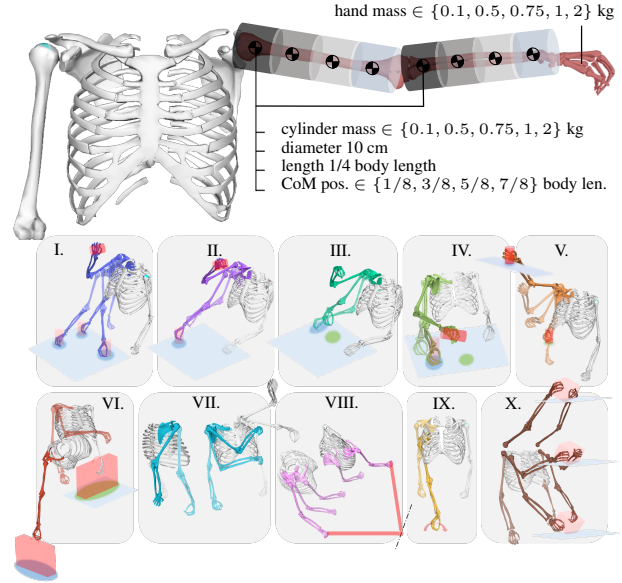
    \vspace{-0.6\baselineskip}
    \caption{Top: Composition of dynamic limb models. 20 cylinders (5 masses, 4 positions) on the humerus; 20 cylinders (5 masses, 4 positions) on the Ulna; 5 hand masses. Bottom: 
   Experimental task set with illustrations outlining trial motion patterns during the manipulation phase and utilised experimental objects.}
    \label{fig: task_set}
    \vspace{-3.5\baselineskip}
\end{figure}

Achieving human-like performance within acceptable device weight is challenged by the limitations of current state-of-the-art actuator technologies. However, prosthetic technologies need not imitate human capabilities but rather be optimised for modern tasks, addressing user concerns. Optimisation and dimension of drive-train topology and components necessitate joint kinematics and kinetics during representative \ac{adl}. Literature on functionally required \ac{RoM} is abundant, with  \cite{Ryu1991_ROM_ADL, Nelson1994_ROM_ADL, Dogan2019_ROM_ADL, Nadeem2022_ROM_ADL, Sardelli2011_ROM_ADL} covering an extensive range of tasks. 

Explicit studies on \ac{adl} joint velocities are comparatively sparse, with \cite{Valevicius2019_ADL_ROM_Agi, Hansson2009_ADL_Agi} focused on small task and joint sets and \cite{MARRAS1993_ADL_Agi} on industrial environments (parts assembly). However, joint velocities can be estimated by differentiation of reported joint angle trajectories of \ac{RoM} studies. 

Functionally required joint kinetics are currently lacking in the literature. The authors are only aware of a singular study \cite{Rosena_ADL_torque} and its derivative \cite{Perry2009_Wrist_Torques_ADL} reporting upper limb joint torques for various \ac{adl} tasks. Object manipulation is a cardinal aspect of \ac{adl}, often resulting in joint torques that surpass those related to limb movement. However, \ac{adl} joint torques induced by objects were not considered in \cite{Rosena_ADL_torque, Perry2009_Wrist_Torques_ADL}, while limb motion-related torques were only examined for limb mass and mass distribution pertaining to the healthy human limb. Consequently, the data reported in \cite{Rosena_ADL_torque, Perry2009_Wrist_Torques_ADL} is of limited usefulness for optimising prosthetic devices exhibiting different dynamic properties. At the same time, neglecting manipulated objects might render resulting mechanisms incapable of completing particular \ac{adl} tasks.

Different methods have been proposed for the consideration of object dynamics in \ac{ID} simulations of \ac{adl} tasks. Muller et al. estimated L5/S1 moments during object transferring tasks by modelling loads as a point mass, whose kinematics were inferred from hand motions \cite{Muller2022_pointMass_torqueestimation}. Akhavanfar et al. investigated differences in predicted spinal loads during bimanual lifting tasks \cite{Akhavanfar2022_2HandSimCompare}. External loads were simulated as unidirectional gravitational loads, additive object mass and inertia to the hand, and wrench-based methods, with experimental object motions tracked by motion capture. Wrench-based procedures were found to be more accurate, producing the least dynamic inconsistencies. Additionally, complex manipulation scenarios where objects are constrained by the environment (e.g., doors, keys, knobs) can be considered by their associated interaction wrench with the hand. Wrench-based methods require accurate object motions $\in \mathbb{R}^6$, which are seldomly reported in \ac{adl} datasets, with \cite{Krebs2021_ADL_KIT_Bimanual, Mandery2015_ADL_KIT_WholeBody} being notable exceptions. However, object kinematics can be accurately calculated from optical hand and wrist marker trajectories \cite{hernethObjectAugmentationAlgorithm2024}, commonly reported in  \ac{adl} databases.

Quality data on task requirements enables data-driven optimisation of mechanism kinematic and dynamic parameters. In \cite{Zarkandi2021_wrist_torque_minim}, the parallel kinematics of a 3 \ac{dof} parallel wrist were optimised for a specified task, effectively reducing required actuator torques. In \cite{Fennel2022, Sha2020}, authors optimised the transparency and topology of assistive devices based on the data reported in \cite{Rosena_ADL_torque}. Damerla et al. \cite{Damerla2022_2Rod_Parallel} created a lightweight prosthetic wrist prototype based on functional requirements in joint \ac{RoM} and velocity. At the same time, the lack of representative data necessitated the oversizing of joint torque capabilities to peak human strength. 

In this study, we provide quantifiable parameters, such as joint \ac{RoM}, velocities, and joint kinetics, necessary for \ac{adl}, that can inform prosthetic device design processes. Joint kinematics were extracted from \cite{Gloumakov2021_ADL_dataset}, reporting optical marker trajectories of 12 subjects performing 24 \ac{adl}. Joint kinetics were computed for a stack of 45 unique limb models per trial, with object-induced torques simulated as described in \cite{hernethObjectAugmentationAlgorithm2024}. \ac{LRM}s trained to torque data of different \ac{adl} tasks, and varying limb mass distributions and object weight permit the computation of necessary peak torques pertaining to limbs and objects of custom dynamic properties. Finally, the data-driven design of a prosthetic wrist is exemplified by optimising joint axis orientations, leveraging a strong correlation between wrist deviation and flexion/extension joint torques.

%% file: model.pdf_tex
\begingroup%
  \makeatletter%
  \providecommand\color[2][]{%
    \errmessage{(Inkscape) Color is used for the text in Inkscape, but the package 'color.sty' is not loaded}%
    \renewcommand\color[2][]{}%
  }%
  \providecommand\transparent[1]{%
    \errmessage{(Inkscape) Transparency is used (non-zero) for the text in Inkscape, but the package 'transparent.sty' is not loaded}%
    \renewcommand\transparent[1]{}%
  }%
  \providecommand\rotatebox[2]{#2}%
  \newcommand*\fsize{\dimexpr\f@size pt\relax}%
  \newcommand*\lineheight[1]{\fontsize{\fsize}{#1\fsize}\selectfont}%
  \ifx\svgwidth\undefined%
    \setlength{\unitlength}{395.02056452bp}%
    \ifx\svgscale\undefined%
      \relax%
    \else%
      \setlength{\unitlength}{\unitlength * \real{\svgscale}}%
    \fi%
  \else%
    \setlength{\unitlength}{\svgwidth}%
  \fi%
  \global\let\svgwidth\undefined%
  \global\let\svgscale\undefined%
  \makeatother%
  \begin{picture}(1,0.88552208)%
    \lineheight{1}%
    \setlength\tabcolsep{0pt}%
    \put(0,0){\includegraphics[width=\unitlength,page=1]{model.pdf}}%
    \put(0.3214518,0.49352467){\makebox(0,0)[lt]{\lineheight{1.25}\smash{\begin{tabular}[t]{l}II.\end{tabular}}}}%
    \put(0.29930954,0.26602757){\makebox(0,0)[lt]{\lineheight{1.25}\smash{\begin{tabular}[t]{l}VII.\end{tabular}}}}%
    \put(0.49777616,0.26579411){\makebox(0,0)[lt]{\lineheight{1.25}\smash{\begin{tabular}[t]{l}VIII.\end{tabular}}}}%
    \put(0.09291972,0.49479224){\makebox(0,0)[lt]{\lineheight{1.25}\smash{\begin{tabular}[t]{l}I.\end{tabular}}}}%
    \put(0.16416297,0.26568713){\makebox(0,0)[lt]{\lineheight{1.25}\smash{\begin{tabular}[t]{l}VI.\end{tabular}}}}%
    \put(0.68765906,0.49352602){\makebox(0,0)[lt]{\lineheight{1.25}\smash{\begin{tabular}[t]{l}IV.\end{tabular}}}}%
    \put(0.84095886,0.49455157){\makebox(0,0)[lt]{\lineheight{1.25}\smash{\begin{tabular}[t]{l}V.\end{tabular}}}}%
    \put(0.65912941,0.26545449){\makebox(0,0)[lt]{\lineheight{1.25}\smash{\begin{tabular}[t]{l}IX.\end{tabular}}}}%
    \put(0.50532419,0.49329027){\makebox(0,0)[lt]{\lineheight{1.25}\smash{\begin{tabular}[t]{l}III.\end{tabular}}}}%
    \put(0.4326057,0.67595987){\makebox(0,0)[lt]{\lineheight{1.25}\smash{\begin{tabular}[t]{l}cylinder mass $\in \{0.1, 0.5, 0.75, 1, 2\}$ kg\\diameter 10 cm \\length 1/4 body length\\CoM pos. $\in \{1/8, 3/8, 5/8, 7/8\}$ body len.\end{tabular}}}}%
    \put(0.9308506,0.85257782){\makebox(0,0)[rt]{\lineheight{1.25}\smash{\begin{tabular}[t]{r}hand mass $\in \{0.1, 0.5, 0.75, 1, 2\}$ kg\end{tabular}}}}%
    \put(0,0){\includegraphics[width=\unitlength,page=2]{model.pdf}}%
    \put(0.76714068,0.26545449){\makebox(0,0)[lt]{\lineheight{1.25}\smash{\begin{tabular}[t]{l}X.\end{tabular}}}}%
    \put(0,0){\includegraphics[width=\unitlength,page=3]{model.pdf}}%
  \end{picture}%
\endgroup%

%% file: chapters/Methodology.tex
\section{Materials and Methods}

\subsection{Activities of daily living dataset}
Joint kinematics and kinetics simulations were based on the \ac{ADLDAT} captured by Gloumakov et al. \cite{Gloumakov2021_ADL_dataset}. The database reports trajectories of 55 optical markers recorded for 12 subjects (6 male, 6 female) completing 18 unimanual and 6 bimanual prescribed \ac{adl} (3 repetitions per task) deemed necessary for independent living. Individual trials distinguished three phases:
\begin{enumerate}
    \item start in a defined pose followed by a reaching motion towards experimental objects
    \item physical object manipulation according to experimental procedures outlined in \cite{Gloumakov2020_Synergies}
    \item returning motion to the starting posture
\end{enumerate}
In this work, \ac{ADLDAT} trials were grouped into ten \ac{adl} tasks based on primitive motions pertaining to each trial's manipulation phase identified in \cite{Gloumakov2020_Synergies, Gloumakov2020_ADL_Decoupling}. Task categories were: I. standing, and II. seated drinking from a cup/mug, III. eating with a fork/spoon, IV. pouring a liquid from a bottle, V. standing overhead picking a tin can, VI. lifting a briefcase from the ground to a tabletop, VII. hygiene (washing axilla and perineal), VIII. opening a door, IX. turning a key/door knob, and X. bimanual placement of a box between 3 shelf heights (low, medium, high). Individual task categories capture different aspects of daily living and the associated requirements for limb joints. The manipulation motions of individual task categories are illustrated in Fig. \ref{fig: task_set}. In the following, limb joints are abbreviated with SR for Glenohumeral internal(+) / external(-) rotation, EF for elbow flexion(+) / extension(-), PS for forearm pronation(+) / supination(-), WF for wrist flexion(+) / extension(-), and WD for wrist Ulnar(+) / Radial(-) deviation. 

\subsection{Joint kinematics simulations}
\label{sec: joint kinematics}
Joint kinematics were computed from optical marker trajectories and subject-specific simulation models with the OpenSim (version 4.4) Python API \cite{Seth2018_OpenSim}. Upper body models were constructed based on the bimanual version of the \ac{dulm} \cite{McFarland2019_MoBL-ARMS}. Limits on the \ac{RoM} of the wrist deviation axis were increased from 10$^{\circ}$ Radial, and 25$^{\circ}$ Ulna deviation to 25$^{\circ}$ and 50$^{\circ}$ respectively. Subject-specific models were generated from individual static trials by scaling the dimension of the \ac{dulm} with the OpenSim scaling tool. Anatomical - maximal and \ac{RMS} marker errors were 1.05 $\pm$ 0.2 cm and 0.66 $\pm$ 0.11 cm, respectively. 

Inverse kinematics were computed with the OpenSim \ac{IK} solver, with joint velocities derived by second-order approximation from low-pass filtered (3rd order, 6 Hz) joint angle trajectories. Trials exhibiting outliers in joint velocities (peaks larger than 1.5 times the inter-quartile range) or maximal \ac{IK} marker errors $>$ 10cm were discarded from our analysis (126 of 864 = 24 experiments * 12 subjects * 3 repetitions). Additionally, slowed joint trajectories were computed by local and proportional deceleration of all joints, such that peak joint velocities never exceeded the 99th percentile velocities of individual joint directions, computed across all full-speed trials. Resulting motions follow the same spatial limb trajectories as full-speed trials. 

\subsection{Joint kinetics simulations}
Torques necessary for the production of low-pass filtered (3rd order, 6 Hz) joint motions (full-speed) of 738 (864-126) individual trials were simulated for:
\begin{enumerate}
    \item varying mass and mass-distribution of the upper arm, forearm, and the hand
    \item wrenches on the hand produced by the manipulation of objects in individual trials 
\end{enumerate}
\ac{ID} simulations were carried out with the Python API of the OpenSim (version 4.4) \ac{ID} tool. 

\subsubsection{Torques for limb motion}
\label{sec: methods, ADL limb torques}
Joint torques were simulated for 45 \ac{dulm}s per subject, with unique dynamic properties of either Humerus, Ulna, or hand bodies. Initially, the mass and inertia of all bodies (except the thorax) were set to 0, while all muscles and passive forces were removed. Subsequently, individual models were constructed:
 \begin{itemize}
     \item Upper arm: modelled from individual cylinders of 10 cm diameter and a mass $\in M= \{0.1, 0.5, 0.75, 1, 2\}$ kg. Cylinder lengths were 1/4 of the subject's Humerus. The cylinder's \ac{CoM} was positioned at intervals $\in \{1/8, 3/8, 5/8, 7/8\}$ of the subject's Humeral length such that the cylinder's axis aligned with the Humerus' (refer to Fig. \ref{fig: task_set}) - 20  models (5 masses at 4 positions).
     \item Forearm: modelled likewise to the upper arm, considering the subject's Ulnar lengths and axes - 20  models (5 masses at 4 positions).
     \item Radius: 0 mass and inertia.
     \item Hand: the subject models' hand mass and inertia properties were scaled to masses $\in M$ kg - 5 models (5 masses).
 \end{itemize}
Trials exhibiting torque peaks larger than 3 times the median torque across all trials of an \ac{adl} (12 subjects, 3 repetitions) were excluded (69 of 738), resulting in 30105 ((736 - 69) trials * 45 models) valid \ac{ID} simulations. \\

\subsubsection{Torques induced by object manipulation}
\label{sec: obj wrench}
Optical markers and force measurements were absent for the objects reported in the \ac{ADLDAT}. Instead, trials were augmented with simulated objects. The interaction between the subject's hands and manipulated objects was modelled as the wrench necessary to move objects along prescribed trajectories. Object kinematics and interaction wrench $\in \mathbb{R}^6$ were estimated from wrist and hand markers with the ObjectAugmentationAlgorithm \cite{hernethObjectAugmentationAlgorithm2024}, grasping information available in \cite{Gloumakov2020_Synergies, Gloumakov2020_ADL_Decoupling, Gloumakov2021_ADL_dataset}, and task-specific object models described in table \ref{tab: tasks_and_objects}. Joint torques pertaining to objects in bi-manual trials of task X were computed as outlined in \cite{Akhavanfar2022_2HandSimCompare}. Task IX (Key/Knob turning) was simulated by applying a static torque around the door knob or key axes to the hand. Tasks III (fork/spoon) and VII were not augmented. Object-related torque terms were isolated by setting all passive forces in the model and mass and inertia of Humerus, Ulna, Radius, and hand bodies to 0. The interaction wrench was applied to the hand by an external loads setup. 

\input{tables/Object_dim}

\subsubsection{Limb and object torque regression}
\label{sec: regr}
The results obtained in sections \ref{sec: methods, ADL limb torques}, and \ref{sec: obj wrench} are now generalized, by fitting \ac{LRM}s to statistical torque quantities of individual joints. Resulting \ac{LRM}s allow the computation of joint torques caused by mass distributions not explicitly simulated. Individual \ac{LRM}s are fitted to joint torque percentiles $p \in \{0, 25, 50, 75, 100\}$ caused by the models of the upper arm, forearm, hand and object models described above. Each \ac{LRM}'s parameter $K_{p,C}$ was fitted to a specific $p$ and a combination $C$ of a joint $J \in$ $\{$SR, EF, PS, WF, WD$\}$, a task $T \in$ $\{$I, II, III, IV, V, VI, VII, VIII, IX, X$\}$, and a body $B \in$ $\{$Humerus, Ulnar, Hand, Object$\}$ according to \eqref{equ: regr}. For Humerus and Ulnar bodies, $\tau_{p,C}$ collected 20 torque values at the $p_{th}$ percentile for each $C$, caused by the 20 cylinders described in section \ref{sec: methods, ADL limb torques}. Their independent variable $X_{cyl}$ was the vector of cylinder \ac{CoM} position $d$ [m] ($d <$ half the cylinder length is not permitted), and cylinder mass $m$ [kg] products \eqref{equ: regr}, with $K_{p,C}$ describing the slope of specific joint torque percentiles with cylinder mass and position. Cylinder positions were $\{1/8, 3/8, 5/8, 7/8\}$ multiplied by the average Humeral/Ulnar length over all subjects. For the hand $\tau_{p,C}$ collected 5 torque values at the $p_{th}$ percentile for each $C$, caused by 5 different hand masses $X_{hand}$. Hand and cylinder masses were $\{0.1, 0.5, 0.75, 1, 2\}$ kg. Objects' $\tau_{p,C}$ collected torque percentile values for each $C$ caused by the manipulation of the task-specific objects described in table \ref{tab: tasks_and_objects}. Each object $X_{obj}$ corresponded to its specific mass or static torque. The hand and objects $K_{p,C}$ describe the slope of percentile joint torques with increasing hand/object mass (or static torque). Torque percentiles $p$ were computed over all trials (12 subjects, 3 repetitions) associated with a $C$. The items 'Cup' and 'Knob' were not considered in the regression analysis since task-specific object torques could not be described by linear models for varying object inertia and static torque signs.

\begin{equation}
\begin{aligned}
& \tau_{p,C} = K_{p,C}X \\
& X_{cyl} = \vert m_1*d_1, m_2*d_2, ... , m_{20}*d_{20} \vert \\
& X_{hand} = \vert m_{1}, m_{2}, ... m_5 \vert \\
& X_{obj} = \vert m_1, m_2, ... , m_n \vert \\
\end{aligned}
\label{equ: regr}
\end{equation}

The superposition principle allows the computation of a composite body's total joint torque requirements by summing the components' torque contributions. Consequently, joint torques for a limb constructed of several cylinders manipulating a specific object can be calculated by adding individual torque contributions. The latter may be computed with the help of the \ac{LRM}s fitted above, allowing the prediction of \ac{adl}-specific 0th, and 100th percentile joint torques for composite, cylinder-based limb mass distributions and task-specific objects. Torques of individual components at the 25th, 50th, and 75th percentile cannot be summed to respective percentiles of the composite limb, since percentile computations need to consider individual data points.  For SR, torque estimations must be calculated as the sum over relevant Humeral, Ulnar, and hand models. In EF, Ulnar and hand models must be considered, whereas only hand models are relevant for PS, WF, and WD. As an example, the maximal wrist flexion torque ($p$=100th, $J$=WF) during the uni-manual lifting task ($T$=VI), for a hand ($B$=Hand) weighing 0.5kg, and a briefcase weighing 1.25kg can be computed according to 

\begin{multline*}
\tau_{100th,C_{Hand}\{WF,VI,Hand\},C_{Brief.}\{WF,VI,Brief.\}} = \\ 
K_{100th, C_{Hand}} * 0.5 + K_{100th, C_{Brief.}} * 1.25 \\
\label{equ: regr applic}
\end{multline*}

\subsection{Wrist axes optimization} \label{sec: methods wrist opti}
\begin{figure*}[tb]
    \centering
    \fontsize{6.5pt}{6.5pt}\selectfont
    \def\svgwidth{1\textwidth}
    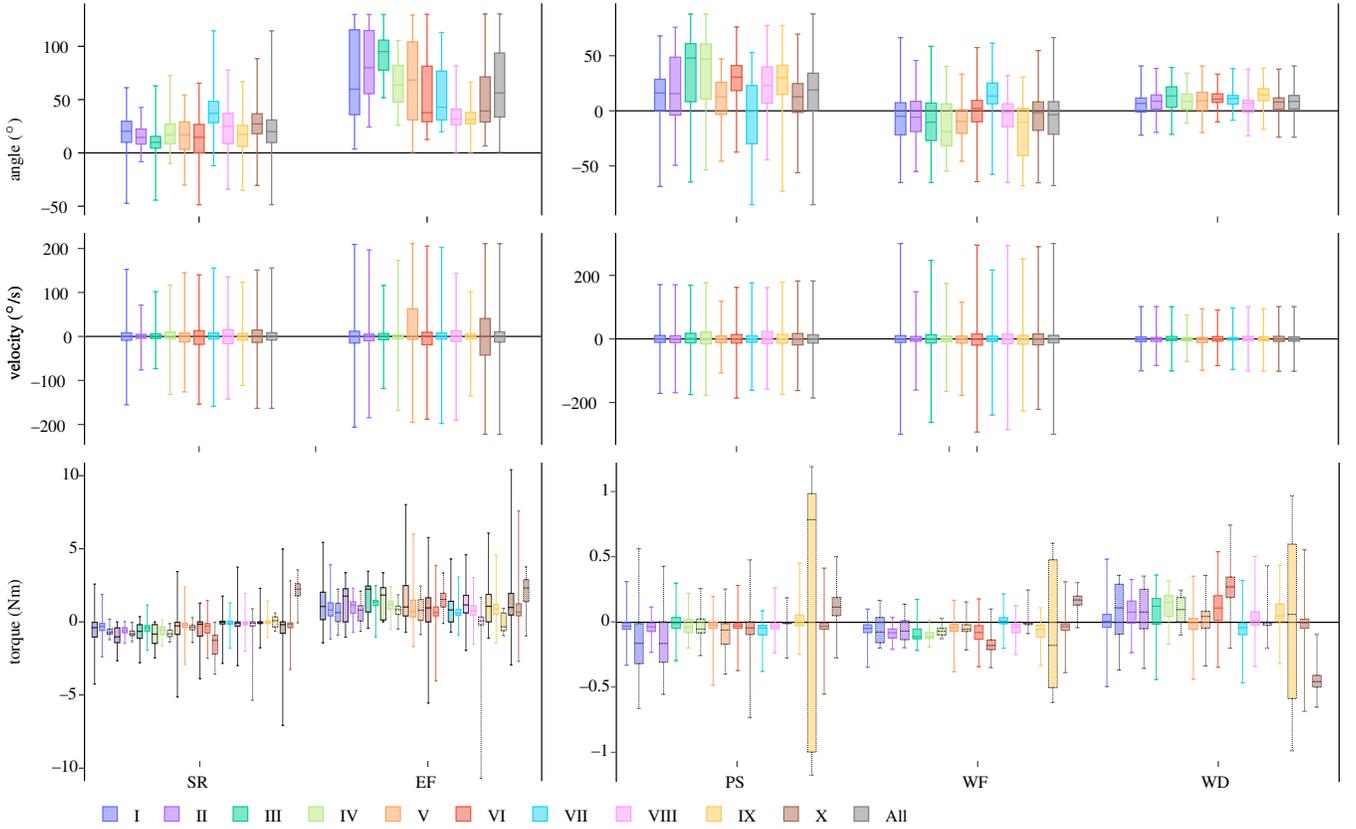
    \vspace{-0.6\baselineskip}
    \caption{Box plots for joint angles ($^{\circ}$) (row 1), joint velocities ($^{\circ}$/s) (row 2), and joint torques (Nm) (row 3). Boxes are specific to joints and tasks, with data concatenated over all subjects and repetitions. Torque Box plots were computed for the Ulna - solid black outline (4 cylinders as depicted in Fig. \ref{fig: task_set} of 0.5kg each), the hand - coloured outline (0.5 kg) and selected task objects $\in$ $\{I $ (Mug 0.5 kg), $II$ (Mug 0.5 kg), $IV$ (Bottle 0.5 kg), $V$ (Briefcase 1kg), $VI$ (Tin can 0.5kg), $IX$ (Key 1.3Nm), $X$ (Box 1kg)$\}$ - dashed black outline.}
    \label{fig: a_vel_m_BOX}
    \vspace{-1.8\baselineskip}
\end{figure*}

This section exemplifies a data-informed design approach for prosthesis joint and actuator design in the example of the wrist. An optimal orientation $\theta \in \{\theta_A, \theta_B\}$ of wrist actuation axes was computed by minimising the maximal sum of simultaneously occurring, absolute actuator powers $P$ in rotated axes \eqref{equ: objective}. Axis index $A$ relates to the WF axis rotated by $\theta_A$, with index $B$ describing the rotated WD axis ($\theta_B$) \eqref{equ: objective}.
 \begin{equation}
\min_{\theta} \quad  \max(|P_A(\theta_A)|) + \max(|P_B(\theta_B)|)
\label{equ: objective}
\end{equation}

$P$ was calculated across all trial's joint velocity and torque pairs, computed in sections \ref{sec: joint kinematics}, and \ref{sec: methods, ADL limb torques}, for a hand weighing 0.5kg. Object induced torques are neglected in this example. Input velocities were capped at WF velocities -300 $^{\circ}$/s $<$ $\nu_{WF}$ $<$ 300 $^{\circ}$/s and WD velocities at -102 $^{\circ}$/s $<$ $\nu_{WD}$ $<$ 102 $^{\circ}$/s, corresponding to $\pm$ 99th percentile joint velocities, deemed sufficient for \ac{adl}. The matrices $\tau$ and $\nu$ contain paired torque and velocity trajectories of WF and WD joints concatenated over all trials.

Wrist joint axes were optimized for four cases:
\begin{itemize}
    \item \textbf{Serial actuation with orthogonal joint axes.} The actuator power for each new axes was calculated by \eqref{equ: Pserial} with axes A and B orthogonal but rotated by $\theta = \theta_A = \theta_B$ from the anatomic orientation.
    \begin{equation}
        P = R(\theta) (\tau \odot \nu)
        \label{equ: Pserial}
    \end{equation}
    $\odot$ denotes the Hadamard product.
    \item \textbf{Serial actuation with non-orthogonal joint axes.} Axes A and B were rotated by distinct angles $\theta_A$ and $\theta_B$. Actuator powers in the rotated system evaluate to \eqref{equ: Pserialnonorth}, with $K^{-1}$ describing the individual rotation of both axes.
    \begin{equation}
        P = K^{-1} (\tau \odot \nu) 
        \label{equ: Pserialnonorth}
    \end{equation}
    \begin{equation}
        K = \begin{Bmatrix}
                        \cos(\theta_A) & \cos(\theta_B)\\
                        \sin(\theta_A) & \sin(\theta_B)
                    \end{Bmatrix}^{-1}
        \label{equ: K}
    \end{equation} 
    \item \textbf{Differential actuation with orthogonal joint axes.} The wrist joint is formed by a differential drive consisting of two actuators and two orthogonal joint axes, rotated by $\theta = \theta_A = \theta_B$ from the anatomic orientation. Joint torques and velocities were calculated from actuator torques and velocities to \eqref{equ: differential transmission torques} and \eqref{equ: differential transmission velocities}, respectively \cite{Wang2013_Differential}. Actuator powers for the rotated mechanism were calculated with \eqref{equ: P differential}.
    \begin{equation}
        P = (T_{\tau}^{-1} R(\theta) \tau) \odot (T_{\nu}^{-1} R(\theta) \nu)
        \label{equ: P differential}
    \end{equation}
    \begin{equation}
        T_{\tau}^{-1} = \begin{Bmatrix}
                            1 & 1\\
                            1 & -1 
                        \end{Bmatrix}^{-1}
        \label{equ: differential transmission torques}
    \end{equation}
    
    \begin{equation}
        T_{\nu}^{-1} = \begin{Bmatrix}
                            0.5 & 0.5\\
                            0.5 & -0.5 
                        \end{Bmatrix}^{-1}
        \label{equ: differential transmission velocities}
    \end{equation} 
    \item \textbf{Differential actuation with non-orthogonal joint axes.} The wrist joint is formed by a differential drive consisting of two actuators and two non-orthogonal joint axes. Joint axes are rotated by distinct angles $\theta_A$ and $\theta_B$ from the anatomic orientation. Consistent with DO, joint torques and velocities were calculated to \eqref{equ: differential transmission torques} and \eqref{equ: differential transmission velocities}. Similarly, actuator powers for the rotated mechanism \eqref{equ: diff P non orhtogonal} were calculated by replacing the rotation matrix $R$ in \eqref{equ: P differential} with the coefficient matrix $K$ \eqref{equ: K}, introduced in the SNO configuration.

    \begin{equation}
        P = (T_{\tau}^{-1} K^{-1} \tau) \odot (T_{\nu}^{-1} K^{-1} \nu)
        \label{equ: diff P non orhtogonal}
    \end{equation}
\end{itemize}

%% file: tables/Object_dim.tex
\begingroup

\setlength{\tabcolsep}{2.5pt} 

\begin{table}[htb]
\caption{Parameters of virtual objects used in each task }
\label{tab: tasks_and_objects}
\begin{tabular}{c|llll}
Task                   & Object    & Osim     & Dimensions {[}mm{]}       & Mass/Torque      \\ \hline
\multirow{2}{*}{I /II} & Cup       & Cylinder & r =43; h=100; handle=40   & 0.1, 0.5, 1 (kg),          \\
                       & Mug       & Cylinder & r =38.5; h=132; handle=30 & 0.1, 0.5, 1 (kg)         \\ \hline
IV                     & Bottle    & Cylinder & r=32; h=213               & 0.1, 0.5, 1, 1.5 (kg)        \\ \hline
V                      & Tin Can   & Cylinder & r=36.5; h=110             & 0.1, 0.5, 1 (kg)         \\ \hline
VI                     & Briefcase & Box      & x=450; y=350; z=110       & 1, 3, 5 (kg)               \\ \hline
VIII                   & Door      & Box      & x=40; y=2032; z=890       & 9, 18, 33 (kg)            \\ \hline
\multirow{2}{*}{IX}    & Key       &          &                           & $\pm$ 0.3, 1.3, 2.3 (Nm)       \\
                       & Knob      &          &                           & 0,3, 1.3, 2.3 (Nm)          \\ \hline
X                      & Box       & Box      & x=340; y=200; z=130       & 0.1, 1, 2, 5 (kg)            \\ 
\end{tabular}
\end{table}

\endgroup

%% file: a_vel_box_INC.pdf_tex
\begingroup%
  \makeatletter%
  \providecommand\color[2][]{%
    \errmessage{(Inkscape) Color is used for the text in Inkscape, but the package 'color.sty' is not loaded}%
    \renewcommand\color[2][]{}%
  }%
  \providecommand\transparent[1]{%
    \errmessage{(Inkscape) Transparency is used (non-zero) for the text in Inkscape, but the package 'transparent.sty' is not loaded}%
    \renewcommand\transparent[1]{}%
  }%
  \providecommand\rotatebox[2]{#2}%
  \newcommand*\fsize{\dimexpr\f@size pt\relax}%
  \newcommand*\lineheight[1]{\fontsize{\fsize}{#1\fsize}\selectfont}%
  \ifx\svgwidth\undefined%
    \setlength{\unitlength}{827.94595073bp}%
    \ifx\svgscale\undefined%
      \relax%
    \else%
      \setlength{\unitlength}{\unitlength * \real{\svgscale}}%
    \fi%
  \else%
    \setlength{\unitlength}{\svgwidth}%
  \fi%
  \global\let\svgwidth\undefined%
  \global\let\svgscale\undefined%
  \makeatother%
  \begin{picture}(1,0.61322484)%
    \lineheight{1}%
    \setlength\tabcolsep{0pt}%
    \put(0.01300405,0.50329839){\rotatebox{90}{\makebox(0,0)[t]{\lineheight{1.25}\smash{\begin{tabular}[t]{c}angle ($^{\circ}$)\end{tabular}}}}}%
    \put(0.01300399,0.36612778){\rotatebox{90}{\makebox(0,0)[t]{\lineheight{1.25}\smash{\begin{tabular}[t]{c}velocity ($^{\circ}$/s)\end{tabular}}}}}%
    \put(0.01300399,0.36612778){\rotatebox{90}{\makebox(0,0)[t]{\lineheight{1.25}\smash{\begin{tabular}[t]{c}velocity ($^{\circ}$/s)\end{tabular}}}}}%
    \put(0.01300404,0.15005901){\rotatebox{90}{\makebox(0,0)[t]{\lineheight{1.25}\smash{\begin{tabular}[t]{c}torque (Nm)\end{tabular}}}}}%
    \put(0.13824786,0.0255459){\makebox(0,0)[lt]{\lineheight{1.25}\smash{\begin{tabular}[t]{l}SR\end{tabular}}}}%
    \put(0.3091533,0.0255459){\makebox(0,0)[lt]{\lineheight{1.25}\smash{\begin{tabular}[t]{l}EF\end{tabular}}}}%
    \put(0.54092643,0.0255459){\makebox(0,0)[lt]{\lineheight{1.25}\smash{\begin{tabular}[t]{l}PS\end{tabular}}}}%
    \put(0.7177809,0.0255459){\makebox(0,0)[lt]{\lineheight{1.25}\smash{\begin{tabular}[t]{l}WF\end{tabular}}}}%
    \put(0.89634221,0.0255459){\makebox(0,0)[lt]{\lineheight{1.25}\smash{\begin{tabular}[t]{l}WD\end{tabular}}}}%
    \put(0.09866804,0.00023302){\makebox(0,0)[lt]{\lineheight{1.25}\smash{\begin{tabular}[t]{l}I\end{tabular}}}}%
    \put(0,0){\includegraphics[width=\unitlength,page=1]{a_vel_box_INC.pdf}}%
    \put(0.14486671,0.00023302){\makebox(0,0)[lt]{\lineheight{1.25}\smash{\begin{tabular}[t]{l}II\end{tabular}}}}%
    \put(0,0){\includegraphics[width=\unitlength,page=2]{a_vel_box_INC.pdf}}%
    \put(0.19620329,0.00023302){\makebox(0,0)[lt]{\lineheight{1.25}\smash{\begin{tabular}[t]{l}III\end{tabular}}}}%
    \put(0,0){\includegraphics[width=\unitlength,page=3]{a_vel_box_INC.pdf}}%
    \put(0.2526636,0.00023302){\makebox(0,0)[lt]{\lineheight{1.25}\smash{\begin{tabular}[t]{l}IV\end{tabular}}}}%
    \put(0,0){\includegraphics[width=\unitlength,page=4]{a_vel_box_INC.pdf}}%
    \put(0.31034116,0.00023302){\makebox(0,0)[lt]{\lineheight{1.25}\smash{\begin{tabular}[t]{l}V\end{tabular}}}}%
    \put(0,0){\includegraphics[width=\unitlength,page=5]{a_vel_box_INC.pdf}}%
    \put(0.36288082,0.00023302){\makebox(0,0)[lt]{\lineheight{1.25}\smash{\begin{tabular}[t]{l}VI\end{tabular}}}}%
    \put(0,0){\includegraphics[width=\unitlength,page=6]{a_vel_box_INC.pdf}}%
    \put(0.42020454,0.00023302){\makebox(0,0)[lt]{\lineheight{1.25}\smash{\begin{tabular}[t]{l}VII\end{tabular}}}}%
    \put(0,0){\includegraphics[width=\unitlength,page=7]{a_vel_box_INC.pdf}}%
    \put(0.48266615,0.00023302){\makebox(0,0)[lt]{\lineheight{1.25}\smash{\begin{tabular}[t]{l}VIII\end{tabular}}}}%
    \put(0,0){\includegraphics[width=\unitlength,page=8]{a_vel_box_INC.pdf}}%
    \put(0.5502515,0.00023302){\makebox(0,0)[lt]{\lineheight{1.25}\smash{\begin{tabular}[t]{l}IX\end{tabular}}}}%
    \put(0,0){\includegraphics[width=\unitlength,page=9]{a_vel_box_INC.pdf}}%
    \put(0.60792906,0.00023302){\makebox(0,0)[lt]{\lineheight{1.25}\smash{\begin{tabular}[t]{l}X\end{tabular}}}}%
    \put(0,0){\includegraphics[width=\unitlength,page=10]{a_vel_box_INC.pdf}}%
    \put(0.66046872,0.00023302){\makebox(0,0)[lt]{\lineheight{1.25}\smash{\begin{tabular}[t]{l}All\end{tabular}}}}%
    \put(0,0){\includegraphics[width=\unitlength,page=11]{a_vel_box_INC.pdf}}%
    \put(0.02932591,0.4559229){\makebox(0,0)[lt]{\lineheight{1.25}\smash{\begin{tabular}[t]{l}−50\end{tabular}}}}%
    \put(0.04571049,0.495844){\makebox(0,0)[lt]{\lineheight{1.25}\smash{\begin{tabular}[t]{l}0\end{tabular}}}}%
    \put(0.03801072,0.53575602){\makebox(0,0)[lt]{\lineheight{1.25}\smash{\begin{tabular}[t]{l}50\end{tabular}}}}%
    \put(0.03031094,0.57566805){\makebox(0,0)[lt]{\lineheight{1.25}\smash{\begin{tabular}[t]{l}100\end{tabular}}}}%
    \put(0,0){\includegraphics[width=\unitlength,page=12]{a_vel_box_INC.pdf}}%
    \put(0.42646777,0.48612416){\makebox(0,0)[lt]{\lineheight{1.25}\smash{\begin{tabular}[t]{l}−50\end{tabular}}}}%
    \put(0.44285235,0.52736779){\makebox(0,0)[lt]{\lineheight{1.25}\smash{\begin{tabular}[t]{l}0\end{tabular}}}}%
    \put(0.43515258,0.56862049){\makebox(0,0)[lt]{\lineheight{1.25}\smash{\begin{tabular}[t]{l}50\end{tabular}}}}%
    \put(0,0){\includegraphics[width=\unitlength,page=13]{a_vel_box_INC.pdf}}%
    \put(0.02162613,0.29253366){\makebox(0,0)[lt]{\lineheight{1.25}\smash{\begin{tabular}[t]{l}−200\end{tabular}}}}%
    \put(0.02162613,0.32547965){\makebox(0,0)[lt]{\lineheight{1.25}\smash{\begin{tabular}[t]{l}−100\end{tabular}}}}%
    \put(0.04571049,0.35842564){\makebox(0,0)[lt]{\lineheight{1.25}\smash{\begin{tabular}[t]{l}0\end{tabular}}}}%
    \put(0.03031094,0.39138068){\makebox(0,0)[lt]{\lineheight{1.25}\smash{\begin{tabular}[t]{l}100\end{tabular}}}}%
    \put(0.03031094,0.42432667){\makebox(0,0)[lt]{\lineheight{1.25}\smash{\begin{tabular}[t]{l}200\end{tabular}}}}%
    \put(0,0){\includegraphics[width=\unitlength,page=14]{a_vel_box_INC.pdf}}%
    \put(0.41876799,0.30896589){\makebox(0,0)[lt]{\lineheight{1.25}\smash{\begin{tabular}[t]{l}−200\end{tabular}}}}%
    \put(0.44285235,0.35656863){\makebox(0,0)[lt]{\lineheight{1.25}\smash{\begin{tabular}[t]{l}0\end{tabular}}}}%
    \put(0.4274528,0.40417137){\makebox(0,0)[lt]{\lineheight{1.25}\smash{\begin{tabular}[t]{l}200\end{tabular}}}}%
    \put(0,0){\includegraphics[width=\unitlength,page=15]{a_vel_box_INC.pdf}}%
    \put(0.03741029,0.03752141){\makebox(0,0)[lt]{\lineheight{1.25}\smash{\begin{tabular}[t]{l}−10\end{tabular}}}}%
    \put(0.04292822,0.0922749){\makebox(0,0)[lt]{\lineheight{1.25}\smash{\begin{tabular}[t]{l}−5\end{tabular}}}}%
    \put(0.05003131,0.14702104){\makebox(0,0)[lt]{\lineheight{1.25}\smash{\begin{tabular}[t]{l}0\end{tabular}}}}%
    \put(0.05003131,0.20176728){\makebox(0,0)[lt]{\lineheight{1.25}\smash{\begin{tabular}[t]{l}5\end{tabular}}}}%
    \put(0.04451329,0.25651342){\makebox(0,0)[lt]{\lineheight{1.25}\smash{\begin{tabular}[t]{l}10\end{tabular}}}}%
    \put(0,0){\includegraphics[width=\unitlength,page=16]{a_vel_box_INC.pdf}}%
    \put(0.44070156,0.04939791){\makebox(0,0)[lt]{\lineheight{1.25}\smash{\begin{tabular}[t]{l}−1\end{tabular}}}}%
    \put(0.43202626,0.0982059){\makebox(0,0)[lt]{\lineheight{1.25}\smash{\begin{tabular}[t]{l}−0.5\end{tabular}}}}%
    \put(0.44780456,0.14702104){\makebox(0,0)[lt]{\lineheight{1.25}\smash{\begin{tabular}[t]{l}0\end{tabular}}}}%
    \put(0.43912926,0.19582903){\makebox(0,0)[lt]{\lineheight{1.25}\smash{\begin{tabular}[t]{l}0.5\end{tabular}}}}%
    \put(0.44780456,0.24464417){\makebox(0,0)[lt]{\lineheight{1.25}\smash{\begin{tabular}[t]{l}1\end{tabular}}}}%
  \end{picture}%
\endgroup%

%% file: chapters/Results.tex
\begin{figure*}[tb]
    \centering
    \fontsize{6.5pt}{6.5pt}\selectfont
    \def\svgwidth{1\textwidth}
    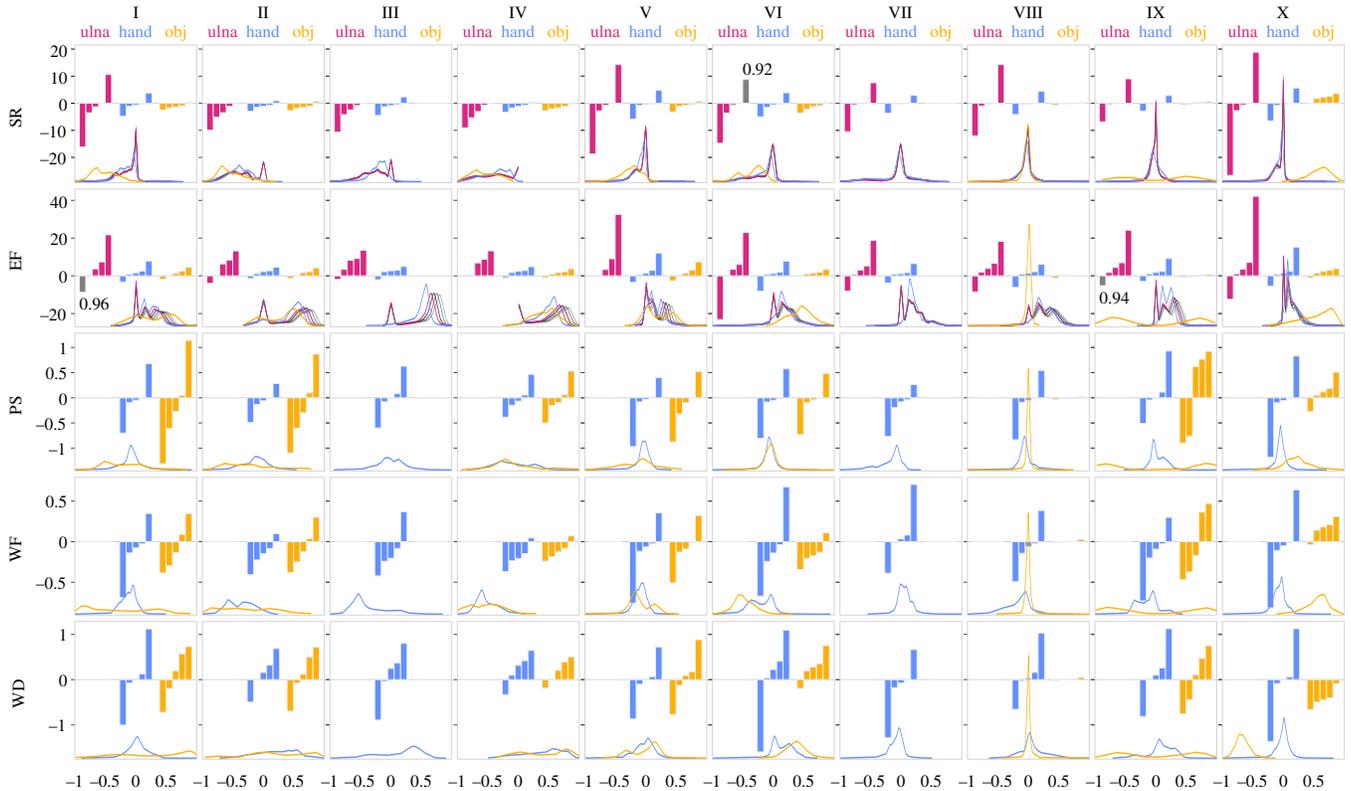
    \vspace{-0.6\baselineskip}
    \caption{Coefficients of linear regression models fitted to percentile joint torques. Colours distinguish forearm (red), hand (blue), and object (yellow) dynamics models. The magnitude of each $K_C$ is indicated by the bar height with percentile models ($p \in \{0, 25, 50, 75, 100\}$) ordered left to right. Grey bars mark models with coefficients of determination R $<$ 0.99. Density plots indicate the torque distributions of normalized joint torques, used in the model fitting. Red: torques produced by a cylinder placed at the most proximal Ulna position (1/8). Grey-scale: positions (3/8, 5/8, 7/8). Blue: torques produced by the hand. Yellow: object induced torques.}
    \label{fig: regressionparam}
    \vspace{-1.8\baselineskip}
\end{figure*}

\section{Results}
This section presents functional joint kinematics and kinetics of human subjects engaging in various \ac{adl}. Results of \ac{IK} simulations pertain to joint angle, velocity and marker error trajectories. \ac{ID} outcomes treat joint torque profiles produced by varying limb mass distributions and torques induced by object manipulation. Sections \ref{sec: joint kinematics}, and \ref{sec: methods, ADL limb torques} inform about exclusion criteria of individual trials from subsequent analysis. 

\subsection{ADL joint kinematics}
\subsubsection{Inverse kinematics errors}
Table \ref{tab: marker_errors} reports maximal and \ac{RMS} errors occurring in \ac{IK} simulations of individual task-types. Maximal errors were the maximally occuring L2 distance (cm) between optical markers of the input dataset and associated virtual markers placed on the \ac{dulm} across all frames of trials related to a task. \ac{RMS} errors were computed accordingly as the average across all trials of a task of maximum \ac{RMS} errors between markers within individual trials. Maximal errors were 6.88 $\pm$ 1.23 cm and \ac{RMS} errors were  0.15 $\pm$ 0.02 cm. \\ 

\input{tables/marker_errors}

\subsubsection{Joint angles} \label{sec: func RoM}
Fig. \ref{fig: a_vel_m_BOX} reports box-plots of \ac{adl}-task specific joint angles, calculated from angle trajectories of each subject's dominant limb. 
\begin{itemize}
    \item Functional SR spanned from 114$^{\circ}$ of internal rotation for hygiene tasks (VII) to 49$^{\circ}$ of external rotation for standing pick and drink (I), eating (III), and uni-manual lifting (VI) tasks. 
    \item Elbow excursions reached 130$^{\circ}$ for tasks I-III, overhead picking (V), VI, and bi-manual pick and place (X). Full elbow extension was observed for tasks I, V, opening a door (VIII), and door key/knob turning (IX).
    \item Forearm pronation angles were maximal for seated eating and pouring tasks (III, IV) 88$^{\circ}$, while supination angles reached 86$^{\circ}$ during the hygiene task.
    \item Wrist flexion peaked around 66$^{\circ}$ for tasks I, III, VI, VII, and X, whereas extension was similar across tasks at 68$^{\circ}$, except for V.
    \item Wrist deviation requirements were consistent across all task types, except IV, VI, and VIII, with 41$^{\circ}$ of Ulnar and 24$^{\circ}$ of Radial deviation.
\end{itemize}

\subsubsection{Joint velocities}
Fig. \ref{fig: a_vel_m_BOX} (second row) shows joint velocity box-plots of individual joints and tasks computed for the slowed motion trajectories described in section \ref{sec: joint kinematics}. 

\begin{itemize}
    \item SR velocities were confined by tasks I, V, VI, VII, and X, ranging from 155 $^{\circ}$/s internal to 164 $^{\circ}$/s external rotation.
    \item EF velocities were maximal for the same tasks as in SR with the addition of task II, reaching 211 $^{\circ}$/s in flexion and 222 $^{\circ}$/s in extension. 
    \item PS velocities peaked around 182 $^{\circ}$/s pronation and 186 $^{\circ}$/s supination for all tasks except V.
    \item WF velocities were highest in tasks I, VI, VIII, and X at 300 $^{\circ}$/s flexion and 300 $^{\circ}$/s extension. As in PS, task V exhibited the lowest speed.
    \item WD velocities were uniform across tasks, at 102 $^{\circ}$/s in ulnar deviation and 102 $^{\circ}$/s in radial deviation.
\end{itemize}

\subsection{ADL joint kinetics} \label{sec: func torque}
Fig. \ref{fig: a_vel_m_BOX} (third row) depicts box plots of individual joint torque contributions of the forearm, hand, and selected objects. Maximal and minimal quantities can be scaled to individual requirements with fitted \ac{LRM}s. Additionally, quartile and median values can be scaled for hand and object models since they do not require model superposition for individual contributions (see section \ref{sec: regr}). Fig. \ref{fig: regressionparam} collects $K_{p,C}$ (height of coloured bars) of individual \ac{LRM}s fitted to combinations of joints (rows), tasks (columns), and dynamic models (colours) at torque percentiles $\in \{0, 25, 50, 75, 100\}$ (bars left to right). Only 3/600 models exhibited coefficients of determination $R$ $<$ 0.99. Torque contributions of the upper arm were negligible relative to those of the forearm, hand, and objects and were therefore excluded from our analysis. Distributions below bars show the density of normalised joint torques used in the computation of depicted $K_{p,C}$ parameters (models fitted to non-normalized data). Numeric $K_{p,C}$ values can be found in the supplementary material. 
\begin{itemize}
    \item SR torques were in external rotation for all but 100th percentile torques, except for object torques in X. Lower percentiles showed greater sensitivity to Ulnar mass and inertia. 
    \item EF torques were primarily positive (flexion), exhibiting an increased sensitivity to forearm mass and mass distribution.
    \item PS, WF, and WD torque percentiles were similarly influenced by hand and object mass in all tasks involving unconstrained objects (I-VI). Task VIII exhibited the least sensitivity to object mass, while Task IX maximised object-related peak and quartile torque requirements in all wrist and forearm joints. Similarly, increased quartile object torques were exhibited by task X.
\end{itemize}

\subsection{Wrist optimisation} \label{sec: wrist opti} 
\begin{figure}[tb]
    \vspace{0.2cm}
    \centering
    \fontsize{7pt}{7pt}\selectfont
    \def\svgwidth{0.49\textwidth}
    \input{Wrist_opti_PCA.pdf_tex}
    \caption{Optimized wrist joint axes and torque principal components in the wrist configuration space. Axes lengths correspond to the required actuator power, with blue and red markers indicating axes A and B, respectively. Purple markers indicate torque samples normalised by $max \sqrt{\tau_{WF}^2 + \tau_{WD}^2}$}.
    \label{fig: WristOptiAxes}
    \vspace{-1.6\baselineskip}
\end{figure}
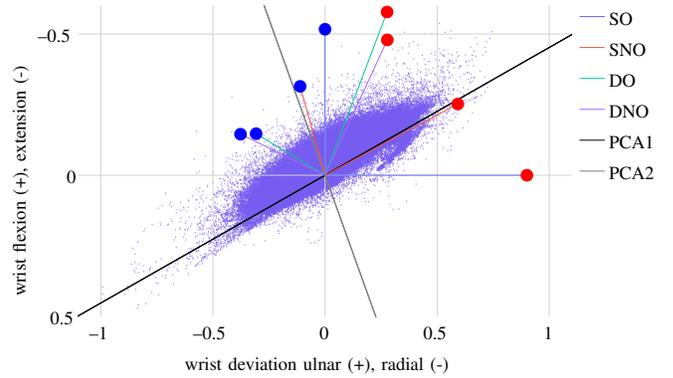

Table \ref{tab: WristOpti} summarises optimised axis orientations and their required \ac{RoM}, torque ($\tau$), velocity ($\nu$), and power (W), as well as percentage changes to the anatomical baseline. Figure \ref{fig: WristOptiAxes} shows torque samples (purple markers) across all trials in the wrist configuration plane, produced by a hand of 0.5kg mass. WF and WD torques exhibited a Pearson correlation coefficient -86 ($p < 0.05$). The principal components of WF and WD torques are depicted as black/grey lines, with the first principal component explaining 95\% of total variance. Optimised A and B axes are displayed as coloured lines, with axis lengths corresponding to the required actuator powers. 

\input{tables/Wrist_opti}
  
In serial actuation with orthogonal joint axes (SO - blue), optimal wrist axes A and B aligned with anatomical WF and WD axes. Serial actuation with non-orthogonal joint axes (SNO - red) resulted in rotated WD and WF axes, coinciding with the first and second torque principal components. Optimisation for differential actuation with orthogonal joint axes (DO - green) and differential actuation with non-orthogonal joint axes (DNO - purple) produced similar A and B axes, wedged centrally between principal components. Optimised joint axes resulted in power reductions from 22\% to 38\%, compared to the baseline. However, joint requirements of optimised axes were susceptible to specific axes orientations, as demonstrated by the varying actuator requirements for nearly overlapping joint axes in DO and DNO.

\input{tables/ProSpec_rom_vel}

%% file: torqueRegr.pdf_tex
\begingroup%
  \makeatletter%
  \providecommand\color[2][]{%
    \errmessage{(Inkscape) Color is used for the text in Inkscape, but the package 'color.sty' is not loaded}%
    \renewcommand\color[2][]{}%
  }%
  \providecommand\transparent[1]{%
    \errmessage{(Inkscape) Transparency is used (non-zero) for the text in Inkscape, but the package 'transparent.sty' is not loaded}%
    \renewcommand\transparent[1]{}%
  }%
  \providecommand\rotatebox[2]{#2}%
  \newcommand*\fsize{\dimexpr\f@size pt\relax}%
  \newcommand*\lineheight[1]{\fontsize{\fsize}{#1\fsize}\selectfont}%
  \ifx\svgwidth\undefined%
    \setlength{\unitlength}{1435.73799674bp}%
    \ifx\svgscale\undefined%
      \relax%
    \else%
      \setlength{\unitlength}{\unitlength * \real{\svgscale}}%
    \fi%
  \else%
    \setlength{\unitlength}{\svgwidth}%
  \fi%
  \global\let\svgwidth\undefined%
  \global\let\svgscale\undefined%
  \makeatother%
  \begin{picture}(1,0.58376811)%
    \lineheight{1}%
    \setlength\tabcolsep{0pt}%
    \put(0,0){\includegraphics[width=\unitlength,page=1]{torqueRegr.pdf}}%
    \put(0.02466586,0.4655426){\makebox(0,0)[lt]{\lineheight{1.25}\smash{\begin{tabular}[t]{l}−20\end{tabular}}}}%
    \put(0.02466586,0.4857639){\makebox(0,0)[lt]{\lineheight{1.25}\smash{\begin{tabular}[t]{l}−10\end{tabular}}}}%
    \put(0.03856069,0.50599044){\makebox(0,0)[lt]{\lineheight{1.25}\smash{\begin{tabular}[t]{l}0\end{tabular}}}}%
    \put(0.03203095,0.52621174){\makebox(0,0)[lt]{\lineheight{1.25}\smash{\begin{tabular}[t]{l}10\end{tabular}}}}%
    \put(0.03203095,0.54643305){\makebox(0,0)[lt]{\lineheight{1.25}\smash{\begin{tabular}[t]{l}20\end{tabular}}}}%
    \put(0,0){\includegraphics[width=\unitlength,page=2]{torqueRegr.pdf}}%
    \put(0.54987873,0.53249968){\makebox(0,0)[lt]{\lineheight{1.25}\smash{\begin{tabular}[t]{l}0.92\end{tabular}}}}%
    \put(0,0){\includegraphics[width=\unitlength,page=3]{torqueRegr.pdf}}%
    \put(0.05385857,0.35710633){\makebox(0,0)[lt]{\lineheight{1.25}\smash{\begin{tabular}[t]{l}0.96\end{tabular}}}}%
    \put(0,0){\includegraphics[width=\unitlength,page=4]{torqueRegr.pdf}}%
    \put(0.02466586,0.34891094){\makebox(0,0)[lt]{\lineheight{1.25}\smash{\begin{tabular}[t]{l}−20\end{tabular}}}}%
    \put(0.03856069,0.37708808){\makebox(0,0)[lt]{\lineheight{1.25}\smash{\begin{tabular}[t]{l}0\end{tabular}}}}%
    \put(0.03203095,0.40526523){\makebox(0,0)[lt]{\lineheight{1.25}\smash{\begin{tabular}[t]{l}20\end{tabular}}}}%
    \put(0.03203095,0.43344237){\makebox(0,0)[lt]{\lineheight{1.25}\smash{\begin{tabular}[t]{l}40\end{tabular}}}}%
    \put(0,0){\includegraphics[width=\unitlength,page=5]{torqueRegr.pdf}}%
    \put(0.81656586,0.36194357){\makebox(0,0)[lt]{\lineheight{1.25}\smash{\begin{tabular}[t]{l}0.94\end{tabular}}}}%
    \put(0,0){\includegraphics[width=\unitlength,page=6]{torqueRegr.pdf}}%
    \put(0.0311956,0.24845736){\makebox(0,0)[lt]{\lineheight{1.25}\smash{\begin{tabular}[t]{l}−1\end{tabular}}}}%
    \put(0.02140099,0.26723692){\makebox(0,0)[lt]{\lineheight{1.25}\smash{\begin{tabular}[t]{l}−0.5\end{tabular}}}}%
    \put(0.03856069,0.28601645){\makebox(0,0)[lt]{\lineheight{1.25}\smash{\begin{tabular}[t]{l}0\end{tabular}}}}%
    \put(0.02876608,0.30479599){\makebox(0,0)[lt]{\lineheight{1.25}\smash{\begin{tabular}[t]{l}0.5\end{tabular}}}}%
    \put(0.03856069,0.32357551){\makebox(0,0)[lt]{\lineheight{1.25}\smash{\begin{tabular}[t]{l}1\end{tabular}}}}%
    \put(0,0){\includegraphics[width=\unitlength,page=7]{torqueRegr.pdf}}%
    \put(0.02140099,0.14829112){\makebox(0,0)[lt]{\lineheight{1.25}\smash{\begin{tabular}[t]{l}−0.5\end{tabular}}}}%
    \put(0.03856069,0.17857345){\makebox(0,0)[lt]{\lineheight{1.25}\smash{\begin{tabular}[t]{l}0\end{tabular}}}}%
    \put(0.02876608,0.20885579){\makebox(0,0)[lt]{\lineheight{1.25}\smash{\begin{tabular}[t]{l}0.5\end{tabular}}}}%
    \put(0,0){\includegraphics[width=\unitlength,page=8]{torqueRegr.pdf}}%
    \put(0.05469124,0.56119837){\color[rgb]{0.8627451,0.14901961,0.49803922}\makebox(0,0)[lt]{\lineheight{1.25}\smash{\begin{tabular}[t]{l}ulna\end{tabular}}}}%
    \put(0.0834393,0.56119837){\color[rgb]{0.39215686,0.56078431,1}\makebox(0,0)[lt]{\lineheight{1.25}\smash{\begin{tabular}[t]{l}hand\end{tabular}}}}%
    \put(0.11798696,0.56119837){\color[rgb]{1,0.69019608,0}\makebox(0,0)[lt]{\lineheight{1.25}\smash{\begin{tabular}[t]{l}obj\end{tabular}}}}%
    \put(0,0){\includegraphics[width=\unitlength,page=9]{torqueRegr.pdf}}%
    \put(0.0311956,0.04174136){\makebox(0,0)[lt]{\lineheight{1.25}\smash{\begin{tabular}[t]{l}−1\end{tabular}}}}%
    \put(0.03856069,0.07554454){\makebox(0,0)[lt]{\lineheight{1.25}\smash{\begin{tabular}[t]{l}0\end{tabular}}}}%
    \put(0.03856069,0.10934251){\makebox(0,0)[lt]{\lineheight{1.25}\smash{\begin{tabular}[t]{l}1\end{tabular}}}}%
    \put(0,0){\includegraphics[width=\unitlength,page=10]{torqueRegr.pdf}}%
    \put(0.15002966,0.56119837){\color[rgb]{0.8627451,0.14901961,0.49803922}\makebox(0,0)[lt]{\lineheight{1.25}\smash{\begin{tabular}[t]{l}ulna\end{tabular}}}}%
    \put(0.1787777,0.56119837){\color[rgb]{0.39215686,0.56078431,1}\makebox(0,0)[lt]{\lineheight{1.25}\smash{\begin{tabular}[t]{l}hand\end{tabular}}}}%
    \put(0.21332539,0.56119837){\color[rgb]{1,0.69019608,0}\makebox(0,0)[lt]{\lineheight{1.25}\smash{\begin{tabular}[t]{l}obj\end{tabular}}}}%
    \put(0,0){\includegraphics[width=\unitlength,page=11]{torqueRegr.pdf}}%
    \put(0.24536807,0.56119837){\color[rgb]{0.8627451,0.14901961,0.49803922}\makebox(0,0)[lt]{\lineheight{1.25}\smash{\begin{tabular}[t]{l}ulna\end{tabular}}}}%
    \put(0.27411615,0.56119837){\color[rgb]{0.39215686,0.56078431,1}\makebox(0,0)[lt]{\lineheight{1.25}\smash{\begin{tabular}[t]{l}hand\end{tabular}}}}%
    \put(0.3086638,0.56119837){\color[rgb]{1,0.69019608,0}\makebox(0,0)[lt]{\lineheight{1.25}\smash{\begin{tabular}[t]{l}obj\end{tabular}}}}%
    \put(0,0){\includegraphics[width=\unitlength,page=12]{torqueRegr.pdf}}%
    \put(0.34070651,0.56119837){\color[rgb]{0.8627451,0.14901961,0.49803922}\makebox(0,0)[lt]{\lineheight{1.25}\smash{\begin{tabular}[t]{l}ulna\end{tabular}}}}%
    \put(0.36945455,0.56119837){\color[rgb]{0.39215686,0.56078431,1}\makebox(0,0)[lt]{\lineheight{1.25}\smash{\begin{tabular}[t]{l}hand\end{tabular}}}}%
    \put(0.40400223,0.56119837){\color[rgb]{1,0.69019608,0}\makebox(0,0)[lt]{\lineheight{1.25}\smash{\begin{tabular}[t]{l}obj\end{tabular}}}}%
    \put(0,0){\includegraphics[width=\unitlength,page=13]{torqueRegr.pdf}}%
    \put(0.43604494,0.56119837){\color[rgb]{0.8627451,0.14901961,0.49803922}\makebox(0,0)[lt]{\lineheight{1.25}\smash{\begin{tabular}[t]{l}ulna\end{tabular}}}}%
    \put(0.46479299,0.56119837){\color[rgb]{0.39215686,0.56078431,1}\makebox(0,0)[lt]{\lineheight{1.25}\smash{\begin{tabular}[t]{l}hand\end{tabular}}}}%
    \put(0.49934066,0.56119837){\color[rgb]{1,0.69019608,0}\makebox(0,0)[lt]{\lineheight{1.25}\smash{\begin{tabular}[t]{l}obj\end{tabular}}}}%
    \put(0,0){\includegraphics[width=\unitlength,page=14]{torqueRegr.pdf}}%
    \put(0.53138332,0.56119837){\color[rgb]{0.8627451,0.14901961,0.49803922}\makebox(0,0)[lt]{\lineheight{1.25}\smash{\begin{tabular}[t]{l}ulna\end{tabular}}}}%
    \put(0.56013142,0.56119837){\color[rgb]{0.39215686,0.56078431,1}\makebox(0,0)[lt]{\lineheight{1.25}\smash{\begin{tabular}[t]{l}hand\end{tabular}}}}%
    \put(0.59467909,0.56119837){\color[rgb]{1,0.69019608,0}\makebox(0,0)[lt]{\lineheight{1.25}\smash{\begin{tabular}[t]{l}obj\end{tabular}}}}%
    \put(0,0){\includegraphics[width=\unitlength,page=15]{torqueRegr.pdf}}%
    \put(0.62672174,0.56119837){\color[rgb]{0.8627451,0.14901961,0.49803922}\makebox(0,0)[lt]{\lineheight{1.25}\smash{\begin{tabular}[t]{l}ulna\end{tabular}}}}%
    \put(0.65546985,0.56119837){\color[rgb]{0.39215686,0.56078431,1}\makebox(0,0)[lt]{\lineheight{1.25}\smash{\begin{tabular}[t]{l}hand\end{tabular}}}}%
    \put(0.69001746,0.56119837){\color[rgb]{1,0.69019608,0}\makebox(0,0)[lt]{\lineheight{1.25}\smash{\begin{tabular}[t]{l}obj\end{tabular}}}}%
    \put(0,0){\includegraphics[width=\unitlength,page=16]{torqueRegr.pdf}}%
    \put(0.72206022,0.56119837){\color[rgb]{0.8627451,0.14901961,0.49803922}\makebox(0,0)[lt]{\lineheight{1.25}\smash{\begin{tabular}[t]{l}ulna\end{tabular}}}}%
    \put(0.75080822,0.56119837){\color[rgb]{0.39215686,0.56078431,1}\makebox(0,0)[lt]{\lineheight{1.25}\smash{\begin{tabular}[t]{l}hand\end{tabular}}}}%
    \put(0.78535594,0.56119837){\color[rgb]{1,0.69019608,0}\makebox(0,0)[lt]{\lineheight{1.25}\smash{\begin{tabular}[t]{l}obj\end{tabular}}}}%
    \put(0,0){\includegraphics[width=\unitlength,page=17]{torqueRegr.pdf}}%
    \put(0.81739864,0.56119837){\color[rgb]{0.8627451,0.14901961,0.49803922}\makebox(0,0)[lt]{\lineheight{1.25}\smash{\begin{tabular}[t]{l}ulna\end{tabular}}}}%
    \put(0.8461467,0.56119837){\color[rgb]{0.39215686,0.56078431,1}\makebox(0,0)[lt]{\lineheight{1.25}\smash{\begin{tabular}[t]{l}hand\end{tabular}}}}%
    \put(0.88069436,0.56119837){\color[rgb]{1,0.69019608,0}\makebox(0,0)[lt]{\lineheight{1.25}\smash{\begin{tabular}[t]{l}obj\end{tabular}}}}%
    \put(0,0){\includegraphics[width=\unitlength,page=18]{torqueRegr.pdf}}%
    \put(0.91273706,0.56119837){\color[rgb]{0.8627451,0.14901961,0.49803922}\makebox(0,0)[lt]{\lineheight{1.25}\smash{\begin{tabular}[t]{l}ulna\end{tabular}}}}%
    \put(0.94148507,0.56119837){\color[rgb]{0.39215686,0.56078431,1}\makebox(0,0)[lt]{\lineheight{1.25}\smash{\begin{tabular}[t]{l}hand\end{tabular}}}}%
    \put(0.97603279,0.56119837){\color[rgb]{1,0.69019608,0}\makebox(0,0)[lt]{\lineheight{1.25}\smash{\begin{tabular}[t]{l}obj\end{tabular}}}}%
    \put(0,0){\includegraphics[width=\unitlength,page=19]{torqueRegr.pdf}}%
    \put(0.01062053,0.49309436){\rotatebox{90}{\makebox(0,0)[lt]{\lineheight{1.25}\smash{\begin{tabular}[t]{l}SR\end{tabular}}}}}%
    \put(0.01061695,0.38590329){\rotatebox{90}{\makebox(0,0)[lt]{\lineheight{1.25}\smash{\begin{tabular}[t]{l}EF\end{tabular}}}}}%
    \put(0.01061339,0.27870073){\rotatebox{90}{\makebox(0,0)[lt]{\lineheight{1.25}\smash{\begin{tabular}[t]{l}PS\end{tabular}}}}}%
    \put(0.01061799,0.16803181){\rotatebox{90}{\makebox(0,0)[lt]{\lineheight{1.25}\smash{\begin{tabular}[t]{l}WF\end{tabular}}}}}%
    \put(0.01264577,0.05909988){\rotatebox{90}{\makebox(0,0)[lt]{\lineheight{1.25}\smash{\begin{tabular}[t]{l}WD\end{tabular}}}}}%
    \put(0.09419894,0.5751213){\makebox(0,0)[lt]{\lineheight{1.25}\smash{\begin{tabular}[t]{l}I\end{tabular}}}}%
    \put(0.18717916,0.5751213){\makebox(0,0)[lt]{\lineheight{1.25}\smash{\begin{tabular}[t]{l}II\end{tabular}}}}%
    \put(0.28016754,0.5751213){\makebox(0,0)[lt]{\lineheight{1.25}\smash{\begin{tabular}[t]{l}III\end{tabular}}}}%
    \put(0.37524123,0.5751213){\makebox(0,0)[lt]{\lineheight{1.25}\smash{\begin{tabular}[t]{l}IV\end{tabular}}}}%
    \put(0.47292133,0.5751213){\makebox(0,0)[lt]{\lineheight{1.25}\smash{\begin{tabular}[t]{l}V\end{tabular}}}}%
    \put(0.56590972,0.5751213){\makebox(0,0)[lt]{\lineheight{1.25}\smash{\begin{tabular}[t]{l}VI\end{tabular}}}}%
    \put(0.65888992,0.5751213){\makebox(0,0)[lt]{\lineheight{1.25}\smash{\begin{tabular}[t]{l}VII\end{tabular}}}}%
    \put(0.75239254,0.5751213){\makebox(0,0)[lt]{\lineheight{1.25}\smash{\begin{tabular}[t]{l}VIII\end{tabular}}}}%
    \put(0.85217365,0.5751213){\makebox(0,0)[lt]{\lineheight{1.25}\smash{\begin{tabular}[t]{l}IX\end{tabular}}}}%
    \put(0.94933136,0.5751213){\makebox(0,0)[lt]{\lineheight{1.25}\smash{\begin{tabular}[t]{l}X\end{tabular}}}}%
    \put(0.04396478,0.00030414){\makebox(0,0)[lt]{\lineheight{1.25}\smash{\begin{tabular}[t]{l}−1\end{tabular}}}}%
    \put(0.06179328,0.00030414){\makebox(0,0)[lt]{\lineheight{1.25}\smash{\begin{tabular}[t]{l}−0.5\end{tabular}}}}%
    \put(0.09295216,0.00030414){\makebox(0,0)[lt]{\lineheight{1.25}\smash{\begin{tabular}[t]{l}0\end{tabular}}}}%
    \put(0.11078061,0.00030414){\makebox(0,0)[lt]{\lineheight{1.25}\smash{\begin{tabular}[t]{l}0.5\end{tabular}}}}%
    \put(0.13930378,0.00030414){\makebox(0,0)[lt]{\lineheight{1.25}\smash{\begin{tabular}[t]{l}−1\end{tabular}}}}%
    \put(0.15713228,0.00030414){\makebox(0,0)[lt]{\lineheight{1.25}\smash{\begin{tabular}[t]{l}−0.5\end{tabular}}}}%
    \put(0.18829116,0.00030414){\makebox(0,0)[lt]{\lineheight{1.25}\smash{\begin{tabular}[t]{l}0\end{tabular}}}}%
    \put(0.20611961,0.00030414){\makebox(0,0)[lt]{\lineheight{1.25}\smash{\begin{tabular}[t]{l}0.5\end{tabular}}}}%
    \put(0.23464278,0.00030414){\makebox(0,0)[lt]{\lineheight{1.25}\smash{\begin{tabular}[t]{l}−1\end{tabular}}}}%
    \put(0.25247122,0.00030414){\makebox(0,0)[lt]{\lineheight{1.25}\smash{\begin{tabular}[t]{l}−0.5\end{tabular}}}}%
    \put(0.28363016,0.00030414){\makebox(0,0)[lt]{\lineheight{1.25}\smash{\begin{tabular}[t]{l}0\end{tabular}}}}%
    \put(0.30145861,0.00030414){\makebox(0,0)[lt]{\lineheight{1.25}\smash{\begin{tabular}[t]{l}0.5\end{tabular}}}}%
    \put(0.32998178,0.00030414){\makebox(0,0)[lt]{\lineheight{1.25}\smash{\begin{tabular}[t]{l}−1\end{tabular}}}}%
    \put(0.34781022,0.00030414){\makebox(0,0)[lt]{\lineheight{1.25}\smash{\begin{tabular}[t]{l}−0.5\end{tabular}}}}%
    \put(0.37896916,0.00030414){\makebox(0,0)[lt]{\lineheight{1.25}\smash{\begin{tabular}[t]{l}0\end{tabular}}}}%
    \put(0.39679755,0.00030414){\makebox(0,0)[lt]{\lineheight{1.25}\smash{\begin{tabular}[t]{l}0.5\end{tabular}}}}%
    \put(0.42532078,0.00030414){\makebox(0,0)[lt]{\lineheight{1.25}\smash{\begin{tabular}[t]{l}−1\end{tabular}}}}%
    \put(0.44314922,0.00030414){\makebox(0,0)[lt]{\lineheight{1.25}\smash{\begin{tabular}[t]{l}−0.5\end{tabular}}}}%
    \put(0.47430811,0.00030414){\makebox(0,0)[lt]{\lineheight{1.25}\smash{\begin{tabular}[t]{l}0\end{tabular}}}}%
    \put(0.49213655,0.00030414){\makebox(0,0)[lt]{\lineheight{1.25}\smash{\begin{tabular}[t]{l}0.5\end{tabular}}}}%
    \put(0.52065977,0.00030414){\makebox(0,0)[lt]{\lineheight{1.25}\smash{\begin{tabular}[t]{l}−1\end{tabular}}}}%
    \put(0.53848822,0.00030414){\makebox(0,0)[lt]{\lineheight{1.25}\smash{\begin{tabular}[t]{l}−0.5\end{tabular}}}}%
    \put(0.56964716,0.00030414){\makebox(0,0)[lt]{\lineheight{1.25}\smash{\begin{tabular}[t]{l}0\end{tabular}}}}%
    \put(0.58747562,0.00030414){\makebox(0,0)[lt]{\lineheight{1.25}\smash{\begin{tabular}[t]{l}0.5\end{tabular}}}}%
    \put(0.61599879,0.00030414){\makebox(0,0)[lt]{\lineheight{1.25}\smash{\begin{tabular}[t]{l}−1\end{tabular}}}}%
    \put(0.63382725,0.00030414){\makebox(0,0)[lt]{\lineheight{1.25}\smash{\begin{tabular}[t]{l}−0.5\end{tabular}}}}%
    \put(0.66498616,0.00030414){\makebox(0,0)[lt]{\lineheight{1.25}\smash{\begin{tabular}[t]{l}0\end{tabular}}}}%
    \put(0.68281467,0.00030414){\makebox(0,0)[lt]{\lineheight{1.25}\smash{\begin{tabular}[t]{l}0.5\end{tabular}}}}%
    \put(0.71133773,0.00030414){\makebox(0,0)[lt]{\lineheight{1.25}\smash{\begin{tabular}[t]{l}−1\end{tabular}}}}%
    \put(0.72916619,0.00030414){\makebox(0,0)[lt]{\lineheight{1.25}\smash{\begin{tabular}[t]{l}−0.5\end{tabular}}}}%
    \put(0.7603251,0.00030414){\makebox(0,0)[lt]{\lineheight{1.25}\smash{\begin{tabular}[t]{l}0\end{tabular}}}}%
    \put(0.77815356,0.00030414){\makebox(0,0)[lt]{\lineheight{1.25}\smash{\begin{tabular}[t]{l}0.5\end{tabular}}}}%
    \put(0.80667678,0.00030414){\makebox(0,0)[lt]{\lineheight{1.25}\smash{\begin{tabular}[t]{l}−1\end{tabular}}}}%
    \put(0.82450524,0.00030414){\makebox(0,0)[lt]{\lineheight{1.25}\smash{\begin{tabular}[t]{l}−0.5\end{tabular}}}}%
    \put(0.85566415,0.00030414){\makebox(0,0)[lt]{\lineheight{1.25}\smash{\begin{tabular}[t]{l}0\end{tabular}}}}%
    \put(0.87349261,0.00030414){\makebox(0,0)[lt]{\lineheight{1.25}\smash{\begin{tabular}[t]{l}0.5\end{tabular}}}}%
    \put(0.90201577,0.00030414){\makebox(0,0)[lt]{\lineheight{1.25}\smash{\begin{tabular}[t]{l}−1\end{tabular}}}}%
    \put(0.91984423,0.00030414){\makebox(0,0)[lt]{\lineheight{1.25}\smash{\begin{tabular}[t]{l}−0.5\end{tabular}}}}%
    \put(0.95100309,0.00030414){\makebox(0,0)[lt]{\lineheight{1.25}\smash{\begin{tabular}[t]{l}0\end{tabular}}}}%
    \put(0.96883155,0.00030414){\makebox(0,0)[lt]{\lineheight{1.25}\smash{\begin{tabular}[t]{l}0.5\end{tabular}}}}%
    \put(0,0){\includegraphics[width=\unitlength,page=20]{torqueRegr.pdf}}%
  \end{picture}%
\endgroup%

%% file: tables/marker_errors.tex
\begingroup

\setlength{\tabcolsep}{4pt} 

\begin{table}[htb]
\caption{Task specific marker errors.}
\vspace{-1\baselineskip}
\label{tab: marker_errors}
\begin{tabular}{l|llllllllll}
Error &    I &   II &  III &   IV &    V &   VI &  VII & VIII &   IX &    X \\ \hline
  Max & 7.41 & 6.71 & 5.67 & 5.20 & 8.08 & 5.76 & 7.38 & 6.25 & 6.82 & 9.55 \\
  RMS & 0.15 & 0.17 & 0.17 & 0.17 & 0.11 & 0.13 & 0.17 & 0.11 & 0.14 & 0.17 \\
\end{tabular}
\vspace{-1.6\baselineskip}
\end{table}

\endgroup

%% file: tables/Wrist_opti.tex

\begingroup

\setlength{\tabcolsep}{3.5pt} 

\begin{table}[tb]
\vspace{0.3cm}
\caption{Wrist range of motion, torque, velocity and power requirements for optimized wrist configurations.}
\vspace{-1\baselineskip}
\label{tab: WristOpti}
\begin{tabular}{lr|rr|rr|rr|rr}

            Axis             & \multicolumn{1}{l}{\begin{tabular}[c]{@{}l@{}}Axis\\ angle [$^{\circ}$]\end{tabular}}  &   \multicolumn{2}{c}{RoM [$^{\circ}$, \%]} &    \multicolumn{2}{c}{$\tau$ [Nm, \%]} &      \multicolumn{2}{c}{$\nu$ [$^{\circ}$/s, \%]} &    \multicolumn{2}{c}{P [W, \%]} \\ \hline

   WD &      0 &   44 &    &  0.8 &  & 273 &  &  2.3 &  \\
   WF &     90 &   69 &    &  0.4 &  & 209 &  &  1.3 &  \\
 SO A &      0 &   44 &     0 &  0.8 &   0 & 273 &   0 &  2.3 &   0 \\
 SO B &     90 &   69 &     0 &  0.4 &   0 & 209 &   0 &  1.3 &   0 \\
SNO A &     23 &   58 &    33 &  0.8 &   8 & 302 &  10 &  1.6 & -29 \\
SNO B &    109 &   66 &    -5 &  0.3 & -21 & 300 &  44 &  0.8 & -35 \\
 DO A &     64 &   82 &    88 &  0.6 & -23 & 426 &  56 &  1.6 & -29 \\
 DO B &    154 &   93 &    35 &  0.2 & -44 & 407 &  95 &  0.9 & -34 \\
DNO A &     60 &   76 &    74 &  0.5 & -29 & 396 &  45 &  1.4 & -38 \\
DNO B &    159 &  101 &    47 &  0.2 & -40 & 443 & 112 &  1.0 & -22 \\

\end{tabular}
\vspace{-1.6\baselineskip}
\end{table}

\endgroup

%% file: wrist_opti_PCA.pdf_tex
\begingroup%
  \makeatletter%
  \providecommand\color[2][]{%
    \errmessage{(Inkscape) Color is used for the text in Inkscape, but the package 'color.sty' is not loaded}%
    \renewcommand\color[2][]{}%
  }%
  \providecommand\transparent[1]{%
    \errmessage{(Inkscape) Transparency is used (non-zero) for the text in Inkscape, but the package 'transparent.sty' is not loaded}%
    \renewcommand\transparent[1]{}%
  }%
  \providecommand\rotatebox[2]{#2}%
  \newcommand*\fsize{\dimexpr\f@size pt\relax}%
  \newcommand*\lineheight[1]{\fontsize{\fsize}{#1\fsize}\selectfont}%
  \ifx\svgwidth\undefined%
    \setlength{\unitlength}{584.96595968bp}%
    \ifx\svgscale\undefined%
      \relax%
    \else%
      \setlength{\unitlength}{\unitlength * \real{\svgscale}}%
    \fi%
  \else%
    \setlength{\unitlength}{\svgwidth}%
  \fi%
  \global\let\svgwidth\undefined%
  \global\let\svgscale\undefined%
  \makeatother%
  \begin{picture}(1,0.56457958)%
    \lineheight{1}%
    \setlength\tabcolsep{0pt}%
    \put(0,0){\includegraphics[width=\unitlength,page=1]{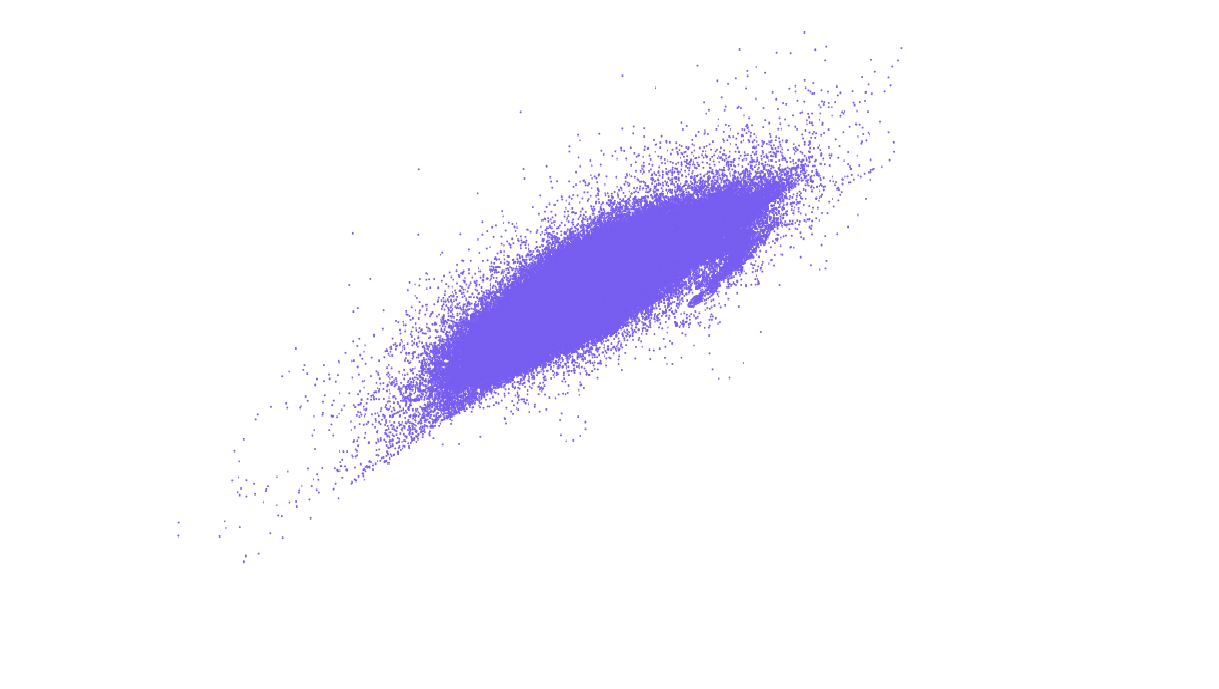}}%
    \put(0.91837095,0.53507998){\makebox(0,0)[lt]{\lineheight{1.25}\smash{\begin{tabular}[t]{l}SO\end{tabular}}}}%
    \put(0,0){\includegraphics[width=\unitlength,page=2]{wrist_opti_PCA.pdf}}%
    \put(0.91837095,0.48789778){\makebox(0,0)[lt]{\lineheight{1.25}\smash{\begin{tabular}[t]{l}SNO\end{tabular}}}}%
    \put(0,0){\includegraphics[width=\unitlength,page=3]{wrist_opti_PCA.pdf}}%
    \put(0.91837095,0.44071555){\makebox(0,0)[lt]{\lineheight{1.25}\smash{\begin{tabular}[t]{l}DO\end{tabular}}}}%
    \put(0,0){\includegraphics[width=\unitlength,page=4]{wrist_opti_PCA.pdf}}%
    \put(0.91837095,0.39353332){\makebox(0,0)[lt]{\lineheight{1.25}\smash{\begin{tabular}[t]{l}DNO\end{tabular}}}}%
    \put(0,0){\includegraphics[width=\unitlength,page=5]{wrist_opti_PCA.pdf}}%
    \put(0.91837095,0.34635108){\makebox(0,0)[lt]{\lineheight{1.25}\smash{\begin{tabular}[t]{l}PCA1\end{tabular}}}}%
    \put(0,0){\includegraphics[width=\unitlength,page=6]{wrist_opti_PCA.pdf}}%
    \put(0.91837095,0.29916885){\makebox(0,0)[lt]{\lineheight{1.25}\smash{\begin{tabular}[t]{l}PCA2\end{tabular}}}}%
    \put(0,0){\includegraphics[width=\unitlength,page=7]{wrist_opti_PCA.pdf}}%
    \put(0.12531559,0.05557557){\makebox(0,0)[lt]{\lineheight{1.25}\smash{\begin{tabular}[t]{l}−1\end{tabular}}}}%
    \put(0.28386321,0.05557557){\makebox(0,0)[lt]{\lineheight{1.25}\smash{\begin{tabular}[t]{l}−0.5\end{tabular}}}}%
    \put(0.47681226,0.05557557){\makebox(0,0)[lt]{\lineheight{1.25}\smash{\begin{tabular}[t]{l}0\end{tabular}}}}%
    \put(0.63535994,0.05557557){\makebox(0,0)[lt]{\lineheight{1.25}\smash{\begin{tabular}[t]{l}0.5\end{tabular}}}}%
    \put(0.81890908,0.05557557){\makebox(0,0)[lt]{\lineheight{1.25}\smash{\begin{tabular}[t]{l}1\end{tabular}}}}%
    \put(0.06589098,0.07852562){\makebox(0,0)[lt]{\lineheight{1.25}\smash{\begin{tabular}[t]{l}0.5\end{tabular}}}}%
    \put(0.09089244,0.29415356){\makebox(0,0)[lt]{\lineheight{1.25}\smash{\begin{tabular}[t]{l}0\end{tabular}}}}%
    \put(0.04709104,0.50978153){\makebox(0,0)[lt]{\lineheight{1.25}\smash{\begin{tabular}[t]{l}−0.5\end{tabular}}}}%
    \put(0.47266705,0.00784078){\color[rgb]{0,0,0}\makebox(0,0)[t]{\lineheight{1.25}\smash{\begin{tabular}[t]{c}wrist deviation ulnar (+), radial (-)\\\end{tabular}}}}%
    \put(0.02844671,0.2977609){\color[rgb]{0,0,0}\rotatebox{90}{\makebox(0,0)[t]{\lineheight{1.25}\smash{\begin{tabular}[t]{c}wrist flexion (+), extension (-) \\\\\end{tabular}}}}}%
    \put(0,0){\includegraphics[width=\unitlength,page=8]{wrist_opti_PCA.pdf}}%
  \end{picture}%
\endgroup%

%% file: tables/ProSpec_rom_vel.tex
\begingroup

\setlength{\tabcolsep}{2.5pt} 

\begin{table}[hb]
\caption{ADL-based, functional prosthesis recommendations for joint RoM and velocities.}
\vspace{-1\baselineskip}
\label{tab: ProSpec}
\begin{tabular}{l|cccccccccc}
              & \multicolumn{2}{c}{SR} & \multicolumn{2}{c}{EF} & \multicolumn{2}{c}{PS} & \multicolumn{2}{c}{WF} & \multicolumn{2}{c}{WD} \\
              & int.    & ext.         & flex.      & ext       & pro.       & sup.      & flex.      & ext.      & Uln.      & Rad.       \\ \hline
RoM ($^{\circ}$)   & 114     & 49 (90)    & 130        & 0         & 88         & 86       & 66         & 68        & 41    & 24  \\
vel 99th ($^{\circ}$/s)     & 155      & 164          & 211         & 222      & 182         & 186       & 300         & 300       & 102        & 102    \\   
vel max ($^{\circ}$/s)     & 267      & 298          & 486         & 381      & 716         & 684       & 425         & 579       & 240        & 260       
\end{tabular}
\end{table}

\endgroup

%% file: chapters/Discussion.tex
\section{Discussion} \label{sec: discussion}

\begin{figure}[b]
    \centering
    \fontsize{6pt}{6pt}\selectfont
    \def\svgwidth{0.45\textwidth}
    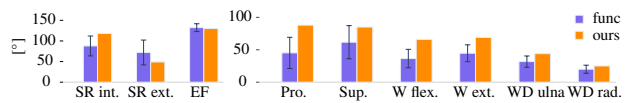
    \vspace{-0.6\baselineskip}
    \caption{Comparison functional joint \ac{RoM} in this and previous studies}
    \label{fig: OtherStudiesComp}
\end{figure}

This study simulated joint kinematics and kinetics of 18 unimanual and 6 bimanual \ac{adl} based on optical marker trajectories of 12 subjects reported in the \ac{ADLDAT}. \cite{Gloumakov2021_ADL_dataset}. The OpenSim Api (4.4) \cite{Seth2018_OpenSim} was employed for \ac{IK} and \ac{ID} computations, with subject-specific models derived by static scaling of the \ac{dulm}. \ac{RoM} constraints of the model's wrist deviation axes were increased since previous research \cite{Gates2015_ROM_ADL, Ryu1991_ROM_ADL, Nelson1994_ROM_ADL, Brigstocke2012_ROM_ADL} showed maximal \ac{adl} joint excursion above the set joint limits. 

Table \ref{tab: ProSpec} collects our recommendations on functionally necessary joint \ac{RoM}, derived as maximally observed joint excursions during the considered \ac{adl}. Fig. \ref{fig: OtherStudiesComp} compares our results with functional joint \ac{RoM} described in the literature. Shoulder internal and external rotation (SR), elbow flexion (EF), forearm supination, and wrist deviation (WD) were within $\pm$ 1 SD of previously reported \ac{adl} joint excursions. Pronation and wrist flexion/extension requirements were found to be more prominent in this study. The absence of additional hygiene tasks, like combing one's hair, caused comparatively low functional internal rotation angles. Consequently, we increased our recommendation for necessary prosthesis internal SR \ac{RoM} from 49$^{\circ}$ to 90$^{\circ}$ (\ref{tab: ProSpec}, bracketed value), based on the results detailed in \cite{Andel2008_ROM_ADL, Magermans2005_ROM_ADL, Raiss2007_ROM_ADL}. The variations in functionally necessary \ac{RoM} reported between studies underline the importance of selecting representative \ac{adl} tasks for analysis. Concurrently, unilateral amputees perform unimanual tasks primarily with the intact limb \cite{chadwellRealityMyoelectricProstheses2016, gambrellOveruseSyndromeUnilateral2008, ostlieMusculoskeletalPainOveruse2011, resnikAdvancedUpperLimb2012}. This includes tasks requiring fine motor control and sensitive tasks such as tooth brushing and combing. On the other hand, bi-manual tasks, such as feeding, require two capable limbs. Therefore, the definition of functional requirements of upper limb prosthetic devices needs to be carefully chosen based on intended usage scenarios (Tasks). 

Joint motions inferred from captured marker trajectories via registration with virtual markers placed on subject models were shown to vary considerably, with small variations in virtual marker placement \cite{Uchida2022_concl_or_illusion}. Consequently, joint kinematics and kinetics computed in this study represent one solution out of a set of equally probable trajectories pertaining to the particular placement of virtual markers on scaled subject models. Additionally, experiments with kinematic inconsistencies, such as gimbal-lock of the ground to thorax joint, associated with extreme velocity spikes, and trials with maximal marker errors above 10cm were excluded from our analysis. However, uncertainties in \ac{IK} reconstructions may persist in reported data.

Joint velocities were calculated by second-order approximation from low-pass filtered joint angle trajectories. Filter cut-off frequencies were chosen to 6 Hz, following the standard settings of OpenSim. Table \ref{tab: ProSpec} reports maximal and 99th percentile velocities of individual joint directions computed across all trials. 99th percentile speeds were substantially exceeded by peak velocities, observed in isolated trials of particular subjects. Therefore, prosthesis design based on maximal joint velocities is not recommended since it would lead to oversized drive trains. In contrast, industrial state-of-the-art prosthesis interfacing schemes do not permit concurrent control of high joint speeds \cite{chadwellRealityMyoelectricProstheses2016}. Consequently, slow-motion trials were computed, allowing a more realistic view of functionally required joint speeds. Spatial limb trajectories in slow trials remained unchanged. At the same time, instantaneous joint velocities were simultaneously and proportionally decelerated such that maximal velocities did not exceed the 99th percentile velocities of respective full-speed joint velocities. Slowed trials exhibited 5 $\pm$ 6 \% longer experiment durations, while required peak velocities were decreased by 53 $\pm$ 14 \%, compared to full-speed trials. Consequently, actuators designed for functional joint speeds of marginally decelerated trials can be dimensioned for half the required peak actuator power. Complementary literature on \ac{adl} joint velocities is scarce, with \cite{Valevicius2019_ADL_ROM_Agi} focused on a limited task set. Functional velocities described for internal (+15\%) and external (+9\%) shoulder rotation, elbow flexion (+28\%), wrist extension (-3\%) and Ulnar wrist deviation (+30\%) compare well with our results. Percentage differences are compared to our recommendations based on 99th percentile velocities. Forearm Pronation was found to be 60\% higher in \cite{Valevicius2019_ADL_ROM_Agi}, while the remaining joints were not reported.

The authors know only one relevant study, \cite{Rosena_ADL_torque}, and its derivative \cite{perryIsotropyUpperLimb2009a}, discussing joint torques produced during a comprehensive set of \ac{adl}. Torques were computed via \ac{ID} for limbs pertaining to healthy human subjects' mass and mass distribution. However, the design of prosthetic devices requires torques produced by the prosthetic mechanisms' specific dynamic properties. Instead, this study simulated joint torques for 45 unique dynamic models per subject and trial. Upper and forearm models consisted of cylinder segments of different masses and \ac{CoM} locations, while the hand was modelled with 5 different weights. Additional to torques necessary for limb motion, interaction forces and moments caused by manipulating objects form a cardinal aspect of \ac{adl}, often resulting in joint torques that surpass those related to limb movement. However, objects were not considered in \cite{Rosena_ADL_torque, Perry2009_Wrist_Torques_ADL}, while \cite{Gloumakov2021_ADLHumanArmMotionData} did not record motions pertaining to individual experimental items. We considered torques induced by object manipulation by applying the wrench necessary for prescribed object motion to the hand. Wrench-based methods were shown to reduce dynamic inconsistencies during \ac{ID} \cite{Muller2022_pointMass_torqueestimation, Akhavanfar2022_2HandSimCompare}. Required object motions $\in \mathbb{R}^6$ were estimated from optical marker trajectories of hand and wrist markers, using our ObjectAugmentationAlgorithm \cite{hernethObjectAugmentationAlgorithm2024}. 

600 \ac{LRM}s were fitted to percentile $p \in \{0, 25, 50, 75, 100\}$ joint torques produced by individual combinations of upper, forearm, hand, and object models, and different \ac{adl} tasks.  Individual \ac{LRM}s exhibited excellent coefficients of determination $R$ = 0.99 $\pm$ 0.01, with only 3/600 models not demonstrating linear relationships between joint torques and limb dynamics. Linearity with cylinder \ac{CoM} position was caused by the dominant influence of gravitational torques, which scale linearly with mass and \ac{CoM} moment arms. Effects proportional to second moments were minor due to generally low \ac{adl} joint velocities and accelerations. \ac{adl} torque \ac{LRM}s permit the prediction of joint toques pertaining to (cylinder-based) mass distributions not explicitly simulated. Superimposing the results of individual \ac{LRM}s allows the designer of mechanisms resembling a human limb, the estimation of peak torques resulting from \ac{adl} limb motion and object manipulation. However, designers need to consider that limb mass, and available control schemes might lead to \ac{adl} joint trajectories distinct from the motions performed by healthy subjects utilised in this study. 

\ac{adl} wrist torques caused by a hand weighing 0.5 kg (without objects) were aligned along an oblique axis, with the first principal component of WF and WD torques explaining 95\% of total variance. This motivated optimising wrist axes orientations for 2 serial (SO, SNO) and 2 differential joint (DO, DNO) configurations, exemplifying data-driven design for upper-limb prosthetics and concomitant reductions in actuator requirements. The optimisation of serial and orthogonal axes produced a wrist configuration coincidental with anatomic joint definitions. Allowing axes to be non-orthogonal aligned individual axes with torque principal components. Optimised DO and DNO axes produced similar configurations, placing torque principal components between optimised axes such that their combined effort could be expanded for peak power requirements. Overall, power requirements of optimised axes orientations distinct from the anatomical configurations lead to reductions in peak power requirements of 22\% to 38\%. The optimisation of an entire prosthetic limb will be addressed in future works. 

Our data provides a valuable basis for informed prosthesis development, keeping in mind the end-user and prevailing over technological limitations when addressing critical flaws in prosthetic device mass and dexterity. 

%% file: otherStudiesComp.pdf_tex
\begingroup%
  \makeatletter%
  \providecommand\color[2][]{%
    \errmessage{(Inkscape) Color is used for the text in Inkscape, but the package 'color.sty' is not loaded}%
    \renewcommand\color[2][]{}%
  }%
  \providecommand\transparent[1]{%
    \errmessage{(Inkscape) Transparency is used (non-zero) for the text in Inkscape, but the package 'transparent.sty' is not loaded}%
    \renewcommand\transparent[1]{}%
  }%
  \providecommand\rotatebox[2]{#2}%
  \newcommand*\fsize{\dimexpr\f@size pt\relax}%
  \newcommand*\lineheight[1]{\fontsize{\fsize}{#1\fsize}\selectfont}%
  \ifx\svgwidth\undefined%
    \setlength{\unitlength}{668.90310693bp}%
    \ifx\svgscale\undefined%
      \relax%
    \else%
      \setlength{\unitlength}{\unitlength * \real{\svgscale}}%
    \fi%
  \else%
    \setlength{\unitlength}{\svgwidth}%
  \fi%
  \global\let\svgwidth\undefined%
  \global\let\svgscale\undefined%
  \makeatother%
  \begin{picture}(1,0.14229372)%
    \lineheight{1}%
    \setlength\tabcolsep{0pt}%
    \put(0,0){\includegraphics[width=\unitlength,page=1]{otherStudiesComp.pdf}}%
    \put(0.04757324,0.02238567){\makebox(0,0)[lt]{\lineheight{1.25}\smash{\begin{tabular}[t]{l}0\end{tabular}}}}%
    \put(0.03860334,0.05667315){\makebox(0,0)[lt]{\lineheight{1.25}\smash{\begin{tabular}[t]{l}50\end{tabular}}}}%
    \put(0.02963343,0.09094941){\makebox(0,0)[lt]{\lineheight{1.25}\smash{\begin{tabular}[t]{l}100\end{tabular}}}}%
    \put(0.02963343,0.12523688){\makebox(0,0)[lt]{\lineheight{1.25}\smash{\begin{tabular}[t]{l}150\end{tabular}}}}%
    \put(0,0){\includegraphics[width=\unitlength,page=2]{otherStudiesComp.pdf}}%
    \put(0.37367925,0.02238567){\makebox(0,0)[lt]{\lineheight{1.25}\smash{\begin{tabular}[t]{l}0\end{tabular}}}}%
    \put(0.36470934,0.07628361){\makebox(0,0)[lt]{\lineheight{1.25}\smash{\begin{tabular}[t]{l}50\end{tabular}}}}%
    \put(0.35573943,0.13017033){\makebox(0,0)[lt]{\lineheight{1.25}\smash{\begin{tabular}[t]{l}100\end{tabular}}}}%
    \put(0.96157941,0.12627491){\makebox(0,0)[lt]{\lineheight{1.25}\smash{\begin{tabular}[t]{l}func\end{tabular}}}}%
    \put(0,0){\includegraphics[width=\unitlength,page=3]{otherStudiesComp.pdf}}%
    \put(0.96157941,0.09958943){\makebox(0,0)[lt]{\lineheight{1.25}\smash{\begin{tabular}[t]{l}ours\end{tabular}}}}%
    \put(0,0){\includegraphics[width=\unitlength,page=4]{otherStudiesComp.pdf}}%
    \put(0.01442773,0.07404538){\rotatebox{90}{\makebox(0,0)[lt]{\lineheight{1.25}\smash{\begin{tabular}[t]{l}[°]\end{tabular}}}}}%
    \put(0.10206672,0.00383674){\makebox(0,0)[lt]{\lineheight{1.25}\smash{\begin{tabular}[t]{l}SR int.\end{tabular}}}}%
    \put(0.18889754,0.00383674){\makebox(0,0)[lt]{\lineheight{1.25}\smash{\begin{tabular}[t]{l}SR ext.\end{tabular}}}}%
    \put(0.29316449,0.00383674){\makebox(0,0)[lt]{\lineheight{1.25}\smash{\begin{tabular}[t]{l}EF\end{tabular}}}}%
    \put(0.44348571,0.00383674){\makebox(0,0)[lt]{\lineheight{1.25}\smash{\begin{tabular}[t]{l}Pro.\end{tabular}}}}%
    \put(0.54049977,0.00383674){\makebox(0,0)[lt]{\lineheight{1.25}\smash{\begin{tabular}[t]{l}Sup.\end{tabular}}}}%
    \put(0.62831629,0.00383674){\makebox(0,0)[lt]{\lineheight{1.25}\smash{\begin{tabular}[t]{l}W flex.\end{tabular}}}}%
    \put(0.72981536,0.00383674){\makebox(0,0)[lt]{\lineheight{1.25}\smash{\begin{tabular}[t]{l}W ext.\end{tabular}}}}%
    \put(0.81960712,0.00383674){\makebox(0,0)[lt]{\lineheight{1.25}\smash{\begin{tabular}[t]{l}WD ulna\end{tabular}}}}%
    \put(0.9198667,0.00383674){\makebox(0,0)[lt]{\lineheight{1.25}\smash{\begin{tabular}[t]{l}WD rad.\end{tabular}}}}%
  \end{picture}%
\endgroup%

%% file: chapters/Acronyms.tex
\begin{acronym}
\acro{ADLDAT}[ADL Dataset]{ADL Human Arm Motion Dataset}
\acro{IQR}[IQR]{Interqartile range}
\acro{FRoM}[FRoM]{Functional Range of Motion}
\acro{RoM}[RoM]{Range of Motion}
\acro{CoM}[CoM]{Center of Mass}
\acro{dof}[DoF]{Degree of Freedom}
\acro{adl}[ADL]{activities of daily living}
\acro{dulm}[MoBL-ARMS DULM]{MoBL-ARMS Dynamic Upper Limb Model}
\acro{tpm}[TPm]{Transhumeral Prosthesis model}
\acro{RMS}[RMS]{Root Mean Square}
\acro{MoCap}[MoCap]{Motion capture}
\acro{CoM}[CoM]{Center of Mass}
\acro{LRM}[LRM]{Linear Regression Model}
\acro{IK}[IK]{Inverse Kinematics}
\acro{ID}[ID]{Inverse Dynamics}
\end{acronym}